\documentclass[10pt,twocolumn,letterpaper]{article}

\usepackage{cvpr}              %
\usepackage[accsupp]{axessibility} %
\usepackage{multirow}
\usepackage{tabularx}
\usepackage{graphicx} %
\usepackage{adjustbox}
\usepackage{listings}
\usepackage{xcolor}

\definecolor{codegreen}{rgb}{0,0.6,0}
\definecolor{codegray}{rgb}{0.5,0.5,0.5}
\definecolor{codepurple}{rgb}{0.58,0,0.82}
\definecolor{backcolour}{rgb}{0.95,0.95,0.92}

\lstdefinestyle{mystyle}{
    backgroundcolor=\color{backcolour},   
    commentstyle=\color{codegreen},
    keywordstyle=\color{magenta},
    numberstyle=\tiny\color{codegray},
    stringstyle=\color{codepurple},
    basicstyle=\ttfamily\footnotesize,
    breakatwhitespace=false,         
    breaklines=true,                 
    captionpos=b,                    
    keepspaces=true,                 
    numbers=left,                    
    numbersep=5pt,                  
    showspaces=false,                
    showstringspaces=false,
    showtabs=false,                  
    tabsize=2
}

\definecolor{cvprblue}{rgb}{0.21,0.49,0.74}
\definecolor{demphcolor}{RGB}{144,144,144}
\newcommand{\demph}[1]{\textcolor{demphcolor}{#1}}

\usepackage[pagebackref,breaklinks,colorlinks,allcolors=cvprblue]{hyperref}

\def\paperID{42715} %
\def\confName{CVPR}
\def\confYear{2026}

\title{SigLino: Efficient Multi-Teacher Distillation for Agglomerative Vision Foundation Models}

\author{
Sofian Chaybouti$^{1,2}$ \textsuperscript{$\dagger$} \hspace{1em} Sanath Narayan$^{1}$ \hspace{1em} Yasser Dahou$^{1}$ \hspace{1em} Phúc H. Lê Khac$^{1}$ \\
Ankit Singh$^{1}$\hspace{2em} Ngoc Dung Huynh$^{1}$\hspace{2em} Wamiq Reyaz Para$^{1}$\hspace{2em} \\  Hilde Kuehne$^{2,3}$\hspace{2em} Hakim Hacid$^{1}$ \vspace{2mm} \\
$^{1}$Technology Innovation Institute, Abu Dhabi, UAE \\
$^{2}$Tuebingen AI Center/University of Tuebingen\\
$^{3}$MIT-IBM Watson AI Lab \\
\footnotesize{Project page: \url{sofianchay.github.io/amoe}}
}

\begin{document}
\maketitle

\begin{figure*}[t]
    \centering
    \includegraphics[width=\textwidth]{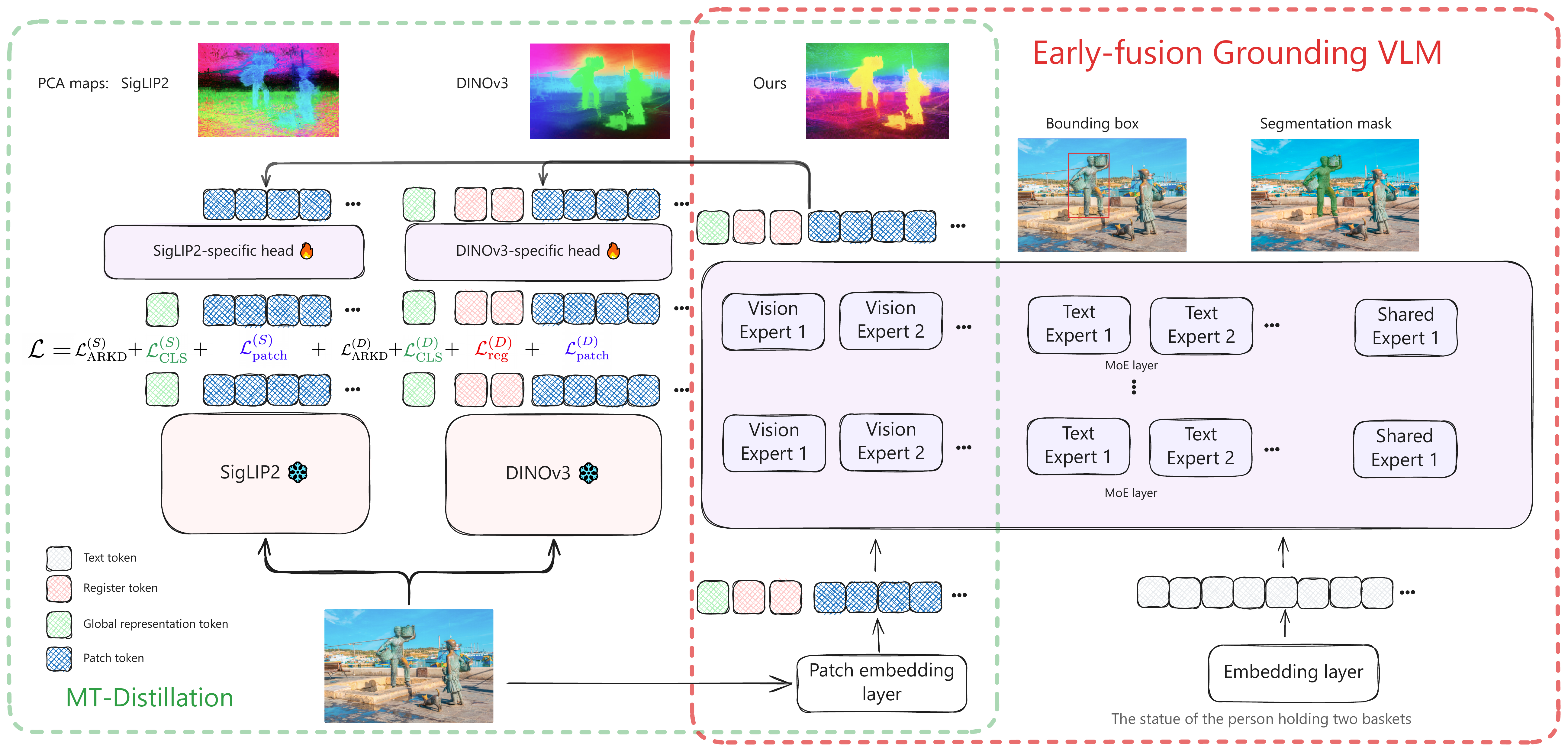}\vspace{-0.25cm}
    \caption{\textbf{SigLino vision foundation model:} A Mixture-of-Experts student is distilled from multiple frozen vision teachers as shown in the multi-teacher distillation stage (on the left). The input image is fed to both teachers (SigLIP2 and DINOv3) and the student to obtain respective patch and global representation embeddings. Additional register tokens are employed in the student model, similar to DINOv3. The student embeddings are then projected to individual teacher embedding spaces via learnable teacher-specific heads. The learning objective includes matching the patch and global (CLS) embeddings of the student with corresponding embeddings of both teachers, in addition to matching the register embeddings with DINOv3 teacher. Moreover, we introduce an asymmetric relational knowledge distillation loss for matching pairwise geometry among samples. The PCA map of the student embeddings (at the top) illustrates the high-quality, dense representations obtained after distillation. Initializing the vision experts of an early-fusion grounding VLM (on the right) from the distilled model improves transfer of teacher representations and achieves strong grounding when training with limited annotations.\vspace{-0.2cm}
}
    \label{fig:teaser_fig}
\end{figure*}

\begin{abstract}

Vision foundation models trained via multi-teacher distillation offer a promising path toward unified visual representations, yet the learning dynamics and data efficiency of such approaches remain underexplored. In this paper, we systematically study multi-teacher distillation for vision foundation models and identify key factors that enable training at lower computational cost. We introduce SigLino, an efficient family of agglomerative vision foundation models that distill knowledge from SigLIP2 and DINOv3 simultaneously into Dense and Mixture-of-Experts students. We show that (1) our Asymmetric Relation-Knowledge Distillation loss preserves the geometric properties of each teacher while enabling effective knowledge transfer, (2) token-balanced batching that packs varying-resolution images into sequences with uniform token budgets stabilizes representation learning across resolutions without sacrificing performance, (3) hierarchical clustering and sampling of training data—typically reserved for self-supervised learning—substantially improves sample efficiency over random sampling for multi-teacher distillation, and (4) the resulting representations transfer effectively to early-fusion Grounding-VLMs, outperforming models trained from scratch. By combining these findings, we curate OpenLVD200M, a 200M-image corpus that demonstrates superior efficiency for multi-teacher distillation. Instantiated in a Mixture-of-Experts, our SigLino-MoE initializes an early-fusion Grounding-VLM that replaces the conventional ViT→LLM stack, demonstrating improved performance compared to a model trained from scratch.  We release OpenLVD200M and five distilled checkpoints comprising MoE and dense variants.

\end{abstract}
    
\section{Introduction}
\label{sec:intro}

Learning universal visual representations that excel across diverse perception tasks remains a fundamental challenge. Recent progress has followed one of two paths: modular vision–language models \cite{bai2025qwen2,yang2025qwen3,lu2024deepseek,wang2024qwen2} that pair a text-aligned vision encoder with a language model, or specialized models trained on single sources of supervision \cite{simeoni2025dinov3, tschannen2025siglip}. While VLMs are effective for instruction-following, they aren't natively multi-modal and often underperform on dense prediction tasks. Single-source foundation models, conversely, excel at their target objective but lack the depth required for general-purpose vision-language understanding.

Recently, an alternative paradigm of agglomerative Vision Foundation Models (VFMs) has emerged, unifying complementary capabilities within a single vision backbone by distilling knowledge from multiple teacher models~\cite{ranzinger2023radio,heinrich2025radiov2}. Although early works in this direction have shown promise, the methodology remains computationally expensive, often requiring a large number of training samples, along with careful consideration for handling varying teacher resolutions and multiple loss functions. A key open question is whether such models can be trained more efficiently in a standardized framework while preserving or even improving their representational quality. To this end, we propose a novel recipe for learning agglomerative VFM, which achieves improved representations with less data, compared to prior works.

We revisit Multi-Teacher (MT) Distillation and identify three critical factors: the quality and distribution of training data, stable multi-resolution training at scale, and the preservation of relational structure geometry. Our investigation yields several key insights. First, we find that uniform coverage of visual concepts through hierarchical clustering clearly outperforms random sampling of equal size, particularly for fine-grained recognition. Second, we show that training on native-resolution images using token-balanced batching and per-image loss normalization stabilizes learning across resolutions, prevents catastrophic forgetting, and improves training efficiency. Third, we demonstrate that preserving the pairwise geometry of teacher embeddings, which we term Asymmetric Relational Knowledge Distillation (ARKD), accelerates learning and improves alignment without sacrificing clustering quality. Finally, we show that a Mixture-of-Experts architecture naturally accommodates complementary teacher signals and enables modality-specific specialization for early-fusion grounding VLMs.

Instantiated with two complementary teachers—SigLIP2~\cite{tschannen2025siglip} for image–text alignment and DINOv3~\cite{simeoni2025dinov3} for dense visual understanding, our student model achieves state-of-the-art performance on global representation benchmarks and competitive results on dense prediction tasks using only 200M curated images. Additionally, we demonstrate that initializing early-fusion grounding VLMs with our distilled vision experts yields strong downstream performance with limited annotation, suggesting more efficient alternatives to classical VLM architectures. Our main contributions are:

\begin{itemize}
    \item We introduce a 200M-image OpenLVD dataset, curated from LAION~\cite{schuhmann2022laion} and DFN~\cite{fang2023data} using hierarchical clustering and balanced sampling~\cite{vo2024automatic}. The OpenLVD dataset facilitates enhanced representation learning during distillation, yielding strong performance on most benchmarks.
    
    \item We optimize the batching technique with token balancing by packing varying-resolution images into sequences with uniform token budgets across batches via FlexAttention~\cite{dong2024flex} and appropriately normalizing the image losses. This achieves stable representation learning across resolutions without sacrificing performance.

    \item We introduce Asymmetric Relation Knowledge Distillation (ARKD) for matching pairwise geometry among samples within a batch via relational knowledge distillation~\cite{park2019relational} to accelerate image-text alignment for DINOv3 ~\cite{jose2025dinov2,zhai2022lit}. Our  ARKD better preserves the clustering properties while improving the learning speed.
            
    \item We show that Mixture-of-Experts (MoE) architecture (Figure~\ref{fig:teaser_fig}) naturally enables early-fusion grounding VLMs via modality-specific experts. Initializing vision experts from our distilled model transfers teacher features, achieving strong grounding performance with limited annotations. Moreover, Gram-Anchoring~\cite{simeoni2025dinov3} preserves dense feature quality during adaptation, preventing the degradation typically observed when learning VLMs.
    
\end{itemize}

\section{Related Work}
\label{sec:rw}

\paragraph{Knowledge Distillation for ViT:}
Knowledge Distillation (KD) has been employed to make large and expensive Vision Transformers (ViT), usually trained on ImageNet~\cite{russakovsky2015imagenet}, lightweight and efficient. The earliest works, such as MiniViT~\cite{zhang2022minivit} and TinyViT~\cite{wu2022tinyvit}, focus on transferring knowledge from large teacher models to small student models. %
Recent works~\cite{chen2022dearkd,yang2024clip,hao2022learning} work on the KD objectives for improving data efficiency. Furthermore, \cite{park2019relational} introduces Relational KD (RKD), which leverages the pairwise relations between samples from the teacher's perspective. In the context of KD for Agglomerative Models trained with Self-Supervised Learning (SSL), we study and improve RKD, demonstrating that it is particularly beneficial for image-text alignment of foundation models aligned with text \textit{a posteriori}, \eg, with the LiT framework~\cite{zhai2022lit}.

\noindent \textbf{Agglomerative Vision Models:} AM-RADIO~\cite{ranzinger2023radio} introduces Agglomerative Vision Models leveraging multi-teacher distillation to build vision foundation models from teachers trained with distinct objectives. SAM-CLIP~\cite{wang2024sam}, Theia~\cite{shang2024theia}, UNIC~\cite{sariyildiz2024unic}, and SAK~\cite{lu2024swiss} are follow-up works. Learning from SAM~\cite{kirillov2023segment}, DFN-CLIP~\cite{fang2023data}, and SigLIP~\cite{zhai2023sigmoid}, RADIOv2.5~\cite{heinrich2025radiov2} significantly improves upon these works by addressing critical challenges, such as resolution mode shift. Here, we refine the multi-teacher distillation recipe to build Agglomerative Vision Models, including MoE variants, focusing on DINOv3~\cite{simeoni2025dinov3} and SigLIP2~\cite{tschannen2025siglip} as teachers. 

\noindent \textbf{Grounding VLM:} Grounding refers to a model's ability to identify and localize specific regions within visual inputs that align with textual descriptions, enabling applications such as referring expression comprehension~\cite{yu2016modeling,kazemzadeh2014referitgame}, \eg, Florence-2~\cite{xiao2024florence} achieves grounding through an image encoder and a sequence-to-sequence architecture. PixelLLM~\cite{xu2024pixel} pretrains on the Localized Narrative dataset~\cite{pont2020connecting} to predict a word with its corresponding position in an image for a powerful foundation for grounding tasks. VisionLLM v2~\cite{wu2024visionllm} introduces “super link” routing tokens with trainable queries that let an LLM condition and invoke task-specific decoders. While these works rely on modular architecture, we aim to avoid this stack of modules through a decoder-only architecture for significant efficiency gains.

\noindent \textbf{Early-Fusion VLM:} The common approach for building VLMs is to connect a pretrained vision encoder to an LLM. While convenient and effective, it is not natively multi-modal. Recent research has focused on connecting pixels and text tokens in an early-fusion fashion to eliminate the need for a vision encoder. Notably,~\cite{shukor2025scaling} shows that native early-fusion multimodal models scale similarly to late-fusion counterparts, for captioning, with sufficient training data. Chameleon~\cite{team2024chameleon} trains an autoregressive early-fusion VLM by encoding images through a trained image tokenizer~\cite{gafni2022make}. VoRA~\cite{wang2025vision} uses LoRA weights to adapt an LLM into an early-fusion VLM, while MonoIntern-VL~\cite{luo2025mono} employs one MLP per modality, and EVE~\cite{diao2024unveiling} uses a unified decoder initialized from an LLM and supervises a module to output a ViT patch representation. Similarly, MoMa~\cite{lin2024moma} shows that MoE with modality-specific experts is optimal for such early-fusion models. Inspired by this, we also use modality-specific experts while using the raw pixels as input. These two reasons justify building an Agglomerative Vision Model through distillation with an MoE Student.

\begin{figure*}[t]
    \centering
    \includegraphics[width=\textwidth]{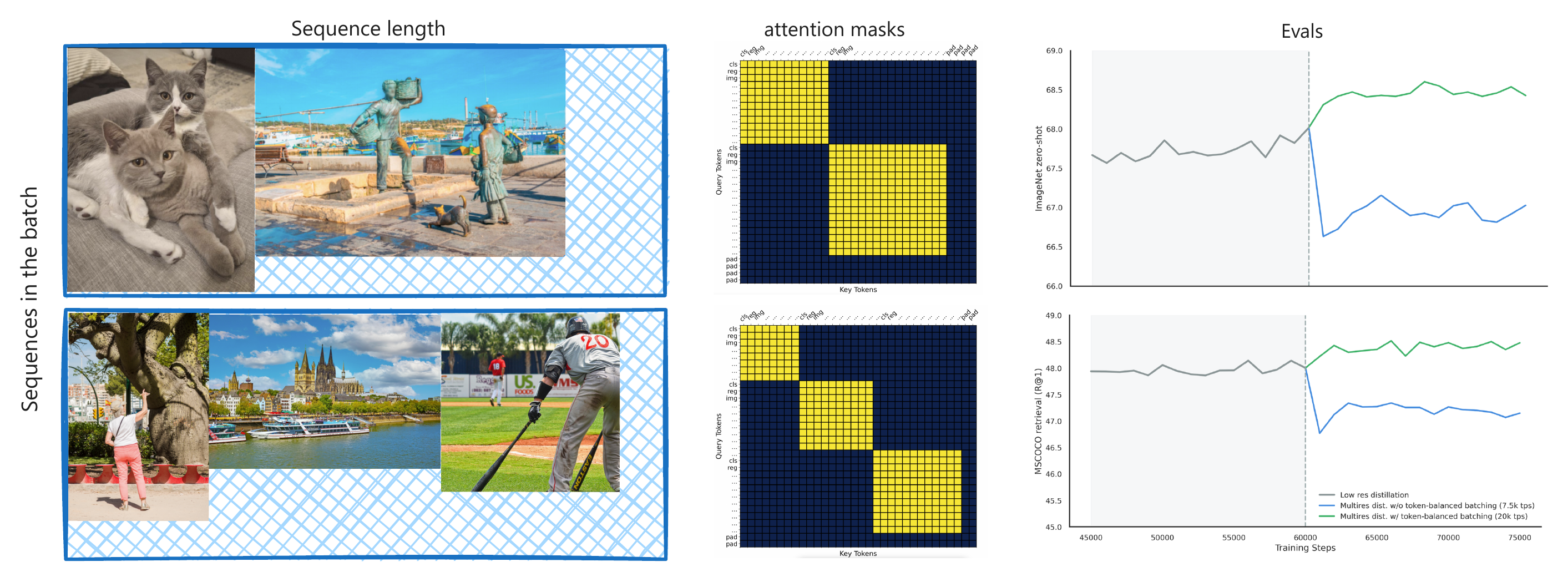}\vspace{-0.25cm}
    \caption{Token-balanced batching: Packing multiple native-resolution images per sequence up to a fixed token budget and applying FlexAttention masks to prevent inter-image attention stabilizes multi-resolution training, prevents low-res forgetting, and improves performance. This strategy also allows for more resource-efficient training with less padding; we go from 7.5k to 20k tokens per second.\vspace{-0.25cm}
}
    \label{fig:token_balanced_batching}
\end{figure*}

\section{Method}
\label{sec:method}

We present our method for building the Agglomerative-MoE Vision Model, later used to initialize an early-fusion grounding VLM with modality-specific experts. Multi‑teacher distillation~\cite{ranzinger2023radio, heinrich2025radiov2} aims to train a single vision encoder that aggregates the strengths of several foundation models. For an input image, the student backbone outputs a global summary token along with patch tokens. Given multiple teachers $\{t_1, \cdots, t_k\}$, per-teacher adaptor heads project these student features into each teacher’s space, the loss aligns global and dense/relational signals from every teacher on the same input. This setting leverages DINOv3’s semantics-rich features and SigLIP2’s language-aligned representations, so that our student inherits both.
We define a “good” MT‑distilled ViT as: (\textit{i}) Global representation quality: strong cluster separation and image–text alignment, reflected in zero‑shot and kNN accuracy. (\textit{ii}) Dense/local quality: semantic fidelity and boundary coherence in patch‑level features, enabling effective linear probes for segmentation. (\textit{iii}) Global–local consistency: the summary token faithfully summarizes, rather than conflicting with, the spatial structure in patch tokens. (\textit{iv}) Teacher fidelity: high per‑teacher feature matching through the adaptor head and ensemble synergy, where the combined supervision outperforms any single teacher, shown in classification ensembling accuracy.

\subsection{Architecture}
We present the MT-distillation, as shown in Figure~\ref{fig:teaser_fig} (left).

\noindent\textbf{Teachers:}
Here, we utilize SigLIP2~\cite{tschannen2025siglip} (ViT-L, Naflex) and DINOv3~\cite{simeoni2025dinov3} (ViT-L) as teachers, as they are two strong native-resolution vision foundation models that provide complementary supervision signals. SigLIP2 is a vision–language encoder contrastively trained with a sigmoid image–text objective and a decoder-style captioning loss. It achieves strong performance on many image-text tasks but suffers from non-separable dense features. In contrast, DINOv3 is trained with self-distillation and Gram-anchoring, designed to preserve extremely high-quality dense features. We aim to learn a student model that simultaneously inherits SigLIP2’s image–text alignment, along with DINOv3’s geometry-patch representations and dense coherence.

\noindent\textbf{Student:}
For the main model, we employ a MoE architecture and two teacher-specific, single-layer MLP projection heads. The backbone tokens are projected into each teacher’s embedding space to supervise patch-level, global features and registers when applicable. We prepend CLS and four register tokens~\cite{darcet2023vision} to the patch tokens, similar to DINOv3. For SigLIP2, the global representation is computed from an attention pooling layer. We adhere to this design and reuse their frozen attention pooling layer, forwarding our SigLIP2-head projected patch features to this module. This avoids re-learning the attention pooling layer and respects how SigLIP2's global summary is represented. Unlike RADIOv2.5~\cite{heinrich2025radiov2}, we use the same projection heads for the patch features and the global image representation.

\subsection{Multi-teacher Distillation Loss}
\label{sec:mt-loss}

\paragraph{Token-balanced batching:} Training on images at native resolution introduces high variance in the number of patch tokens per sample (e.g., $256\times256$ images yield $\sim 256$ patches while $768\times768$ yield $\sim 2{,}304$ ). Naively batching fixed numbers of images per rank leads to dramatically unbalanced token counts across ranks, which destabilizes optimization and causes high-norm gradients.

We address this through \emph{token-balanced batching}, where multiple images are packed~\cite{dehghani2023patch} into sequences up to a maximum context length $C_{\max}$ and avoid inter-image self-attention via FlexAttention~\cite{dong2024flex}. This yields approximately uniform token budgets per rank, but introduces a new challenge: each packed sequence may contain a different number of images, and losses must be normalized correctly to ensure stable, unbiased gradients across images and ranks. Figure~\ref{fig:token_balanced_batching} illustrates this concept. On the right, we see that token-balanced batching avoids forgetting image representations at low resolutions; even better, it improves them.

\noindent \textbf{Notation:} Let $\mathcal{T}$ denote the set of teachers and $t\in\mathcal{T}$ a fixed teacher. Training proceeds over $R$ distributed ranks,
 where rank $r\in\{1,\dots,R\}$. Let $J_r$ be the number of packed sequences and $I_r^{(j)}$ the number of images in sequence $j\in\{1,\dots,J_r\}$. The total number of images in the global batch is $B_{\mathrm{global}}=\sum_{r=1}^R\sum_{j=1}^{J_r}I_r^{(j)}$. Let $N_{r,j,i}$ denote the number of patch tokens for a particular image indexed by $(r,j,i)$ (rank $r$, sequence $j$, image $i\in\{1,\dots,I_r^{(j)}\}$).
For teacher $t$ and image $q$ (with $q {=} (r,j,i)$ for convenience):
\begin{itemize}[leftmargin=*, itemsep=2pt, parsep=0pt]
    \item $z^{(t,s)}_q\in\mathbb{R}^{d_t}$ is the teacher \emph{summary} embedding, $\hat z^{(t,s)}_q\in\mathbb{R}^{d_t}$ the projected student summary.
    \item $\{z^{(t,p)}_{q,\ell}\}_{\ell=1}^{N_{q}}\subset\mathbb{R}^{d_t}$ are teacher \emph{patch} embeddings, $\{\hat z^{(t,p)}_{q,\ell}\}_{\ell=1}^{N_{q}}\subset\mathbb{R}^{d_t}$ the projected student patches.
\end{itemize}
We denote similarity as $\cos(u,v){=}\langle u,v\rangle/(\lVert u\rVert_2\lVert v\rVert_2)$. For DINOv3, let $K$ be the number of registers, with $z^{(t,reg)}_{q,k}$ and $\hat z^{(t,reg)}_{q,k}$ denoting teacher and student register embeddings.

\noindent \textbf{Per-image losses with token-based normalization:} Following RADIOv2.5~\cite{heinrich2025radiov2}, we align student and teacher global (summary, registers) and local (patch-wise) representations through teacher-specific projection heads. Moreover, to prevent high-resolution images from dominating the gradient, we normalize patch and register losses by the number of tokens \emph{per image} before aggregating globally. For image $q{=}(r,j,i)$ and teacher $t$, the per-image losses are:
\begin{equation}
\label{eq:per-image-sum}
\mathcal{L}^{(t)}_{\mathrm{CLS}}(q)
=
1-\cos\bigl(z^{(t,s)}_{q},\hat z^{(t,s)}_{q}\bigr),
\end{equation}
\begin{equation}
\label{eq:per-image-patch}
\mathcal{L}^{(t)}_{\mathrm{patch}}(q)
=
\frac{1}{N_{q}}\sum_{\ell=1}^{N_{q}}\!\lVert z^{(t,p)}_{q,\ell}-\hat z^{(t,p)}_{q,\ell}\rVert_2^2,
\end{equation}
\begin{equation}
\label{eq:per-image-reg}
\mathcal{L}^{(t)}_{\mathrm{reg}}(q)
=
\mathbf{1}_{t=\text{DINO}}\,\frac{1}{K}\sum_{k=1}^{K}\!\bigl\lVert z^{(t,reg)}_{q,k}-\hat z^{(t,reg)}_{q,k}\bigr\rVert_2^2.
\end{equation}
The combined per-image loss for teacher $t$ is
\begin{equation}
\label{eq:per-image-combined}
\mathcal{L}^{(t)}(q)
=
\mathcal{L}^{(t)}_{\mathrm{CLS}}(q)
+
\mathcal{L}^{(t)}_{\mathrm{patch}}(q)
+
\mathcal{L}^{(t)}_{\mathrm{reg}}(q).
\end{equation}
\paragraph{Global batch aggregation:}
To ensure unbiased gradients, we average the per-image losses across all images in the global batch, regardless of how they are packed:
\begin{equation}
\label{eq:teacher-loss-global}
\mathcal{L}^{(t)}_{\mathrm{global}}
=
\frac{1}{B_{\mathrm{global}}}
\sum_{r=1}^{R}\sum_{j=1}^{J_r}\sum_{i=1}^{I_r^{(j)}}
\mathcal{L}^{(t)}(q).
\end{equation}
The final multi-teacher objective sums over all teachers:
\begin{equation}
\label{eq:final-loss}
\mathcal{L}_{\mathrm{total}}
=
\sum_{t\in\mathcal{T}}\mathcal{L}^{(t)}_{\mathrm{global}}.
\end{equation}
 
This ensures: (\textit{i}) images contribute equally to the loss regardless of resolution, (\textit{ii}) token counts are balanced across ranks for stable throughput, and (\textit{iii}) gradients remain well-scaled across the heterogeneous resolution distribution.

\begin{figure}[t]
    \centering
    \includegraphics[width=\columnwidth]{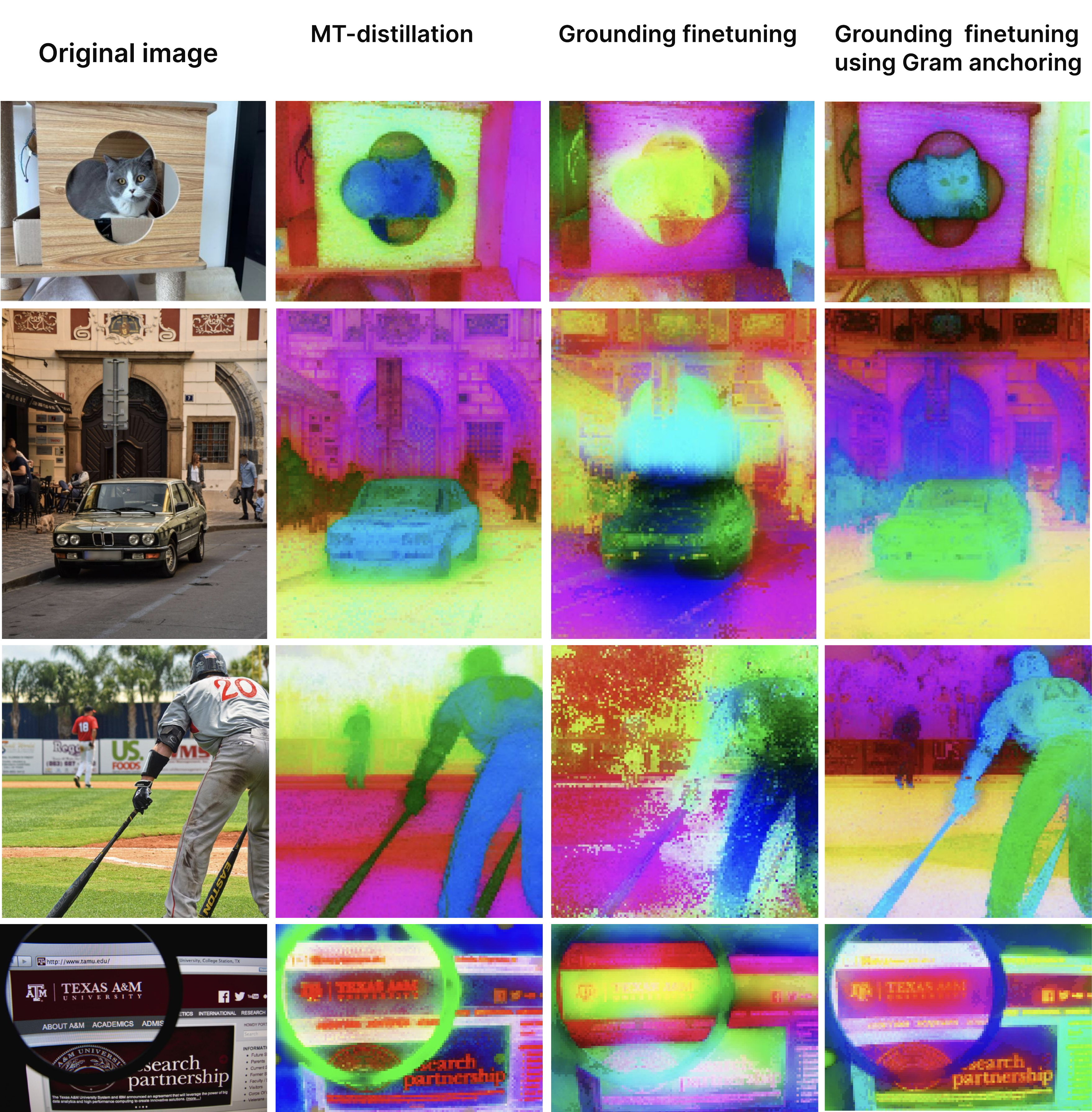}\vspace{-0.25cm}
    \caption{
    PCA maps of patch features across training stages. Columns (left→right): original image; PCA of MT‑distilled features; PCA after grounding fine‑tuning; PCA after grounding fine‑tuning with Gram‑anchoring. Naive fine‑tuning degrades spatial structure. Adding Gram-anchoring preserves spatial coherence from SigLino while still adapting the representation.\vspace{-0.25cm} %
}
    \label{fig:gram-anchoring}
\end{figure}

\noindent\textbf{Teacher-loss balancing via PHI-S:} PHI-S~\cite{ranzinger2024phi} (PCA–Hadamard Isotropic Standardization) is a normalization technique for label-free multi-teacher distillation that equalizes the statistical scales of diverse teacher feature distributions and distributes per-channel variance equally before the student learns to match them. The different teachers have very different variances and means, so MSE/Smooth-L1 implicitly overweights high-variance teachers and channels. PHI-S normalizes each teacher target with an invertible linear mapping during training, and then inverts it at inference so the student still outputs features in the teacher’s original space. Roughly speaking, PHI-S rotates the features via an invertible matrix built from Hadamard Matrices and second-order moments estimation. For each type of feature and each teacher, we learn a PHI-S transform on 3 million samples from our training data. However, for DINOv3, we observed that the PHI-S transform of the second register cannot be accurately estimated, as it exhibits multiple modes. Hence, when estimating a mean and a covariance matrix, it is representative of between-mode statistics, and the features cannot be well centered and scaled. In practice, we observe that it leads to high-norm gradients and dramatically slows down learning. Further analysis on these elements is provided in the supplement. For simplicity, we do not apply the PHI-S transform to any register during MT-distillation.

\noindent\textbf{Asymmetric Relational Knowledge Distillation:} We investigate whether augmenting one-to-one global representation matching with a relational loss, inspired by relational knowledge distillation~\cite{park2019relational} (RKD), is beneficial. Instead of only aligning teacher and student embeddings per sample, we also match the pairwise geometry among samples within a batch. In practice, we observe that it is very beneficial for image-text alignment with DINOv3, while the gains are marginal for SigLIP2. We provide two explanations for this: (1) DINOv3 is aligned with text only \textit{a posteriori} through the LiT procedure~\cite{zhai2022lit}, resulting in lower ground-truth image-text similarity scales ($0.2$ vs. $0.9$ for SigLIP2). (2) The relational loss does not decrease with the global representation loss for DINOv3, serving as a regularization term that enforces correct distances between samples.
However, while beneficial for image-text alignment, we observe that RKD harms kNN performances. We hypothesize this is due to the loss aggressively pushing or attracting samples when they should be relatively far apart in the embedding space. We propose a simple fix: making RKD asymmetric (ARKD) by bringing two samples closer or pushing them only if they are close/far in teacher space. We use the intra-batch median of embedding distances in teacher space as the decision boundary.
Mathematically, let $t_i{=}z^{(t,s)}_{i}$ and $s_i{=}\hat z^{(t,s)}_{i}$ be teacher and student \emph{summary} embeddings. We define $D^T_{ij}=d(t_i, t_j)$, $D^S_{ij}=d(s_i,s_j)$, where $d(x,y)=\lVert x-y\rVert_2$, the teacher scale $\bar D^T=\tfrac{1}{B_{global}(B_{global}-1)}\sum_{i\neq j}D^T_{ij}$, and normalized distances $\hat D^T_{ij}=D^T_{ij}/\bar D^T$, $\hat D^S_{ij}=D^S_{ij}/\bar D^T$ with $m=\operatorname{median}(\{\hat D^T_{ij}\}_{i\neq j})$. Using one-sided errors with binary split:
$\text{shrink}_{ij}=\max\{\hat D^S_{ij}-\hat D^T_{ij},0\}$,
$\text{expand}_{ij}=\max\{\hat D^T_{ij}-\hat D^S_{ij},0\}$,
$w_{\text{shrink},ij}=\mathbf{1}\{\hat D^T_{ij}<m\}$,
$w_{\text{expand},ij}=1-w_{\text{shrink},ij}$.
With the smooth-L1 function $h(\cdot)$, the loss is:

\begin{equation}
\label{eq:rkd-compact-short}
\begin{aligned}
\mathcal{L}^{(t)}_{\mathrm{ARKD}}
&= \frac{1}{B_{global}(B_{global}-1)}\sum_{i\neq j}\\
&\quad\Bigl(w_{\text{expand},ij}\,h(\text{expand}_{ij})\\
&\quad\quad+w_{\text{shrink},ij}\,h(\text{shrink}_{ij})\Bigr).
\end{aligned}
\end{equation}

\noindent The per-teacher objective is: $\mathcal{L}^{(t)}=\mathcal{L}^{(t)}_{\mathrm{global}}+\mathcal{L}^{(t)}_{\mathrm{ARKD}}$.

\subsection{Curating OpenLVD200M}
We utilize the hierarchical clustering and sampling technique, introduced by~\cite{vo2024automatic}, to mitigate long-tail biases in web-scraped datasets. This has been demonstrated to flatten concept distributions and enhance SSL performances, both theoretically and in practice, and has been successfully applied to train DINOv3 (LVD-1.7B, curated from 17B original samples). We introduce OpenLVD200M, constructed from a 2.3B-image blend of DFN and LAION. We make a few efficiency adjustments to the original algorithm, allowing it to run on 12 A100 nodes instead of the estimated 45 nodes with the original algorithm. These are fully detailed in the supplementary material. Concretely, we embed images with DINOv3 ViT-B encoder and (\textit{i}) uniformly subsample 1B images, (\textit{ii}) run a 4-level hierarchical clustering with 20M, 500k, 50k, and 20k centroids, (\textit{iii}) assign the remaining 1.7B images to the 20M level-1 centroids, and (\textit{iv}) perform hierarchical sampling to obtain a balanced 200M-image subset. This curation yields broader, more uniform concept coverage that we hypothesize and demonstrate experimentally to be especially beneficial for MT-distillation.

\subsection{High-resolution Training}

We adopt a two-stage recipe for high-resolution distillation. In stage~1, we distill on OpenLVD up to \(256\times256\) to rapidly learn strong global and dense representations. In stage~2, we post-train for high resolution (up to \(768\times768\) on 13M images (11.5M from SAM~\cite{kirillov2023segment} and 1.5M web-scraped). Naively using this pool causes a distribution shift, resulting in the forgetting of low-resolution global features and degraded performance. Our \textit{token-balanced batching} and per-image token-normalized losses (\S\ref{sec:mt-loss}) are critical to making this stage stable and effective, ensuring that high-resolution images do not dominate gradients while maintaining uniform computational load across ranks. We train on a multi-resolution blend that preserves the low-resolution distribution while introducing high-resolution content: we reintroduce OpenLVD at \(256\times256\), include the images with natural sizes between \(256\times256\) and \(384\times384\), and add the high-resolution pool down-sampled to \(256\times256\) and \(512\times512\), maintaining the natural data distribution.

\subsection{Early-fusion Grounding VLM Initialization}

Whether good image representations can arise solely from next-token prediction in early fusion remains underexplored~\cite{shukor2025scaling}. Here, we explore early-fusion grounding VLM using MoE architecture with modality-specific and shared experts, and analyze the effects of initializing vision experts via distillation. Particularly, we initialize a lightweight 0.2B-1B 12-layer MoE model with 28 (six active) per-modality experts and 8 shared experts (two active), initializing the vision experts using the AMoE distilled student.

Our distillation recipe leverages SigLIP2's vision-language alignment and DINOv3's dense features, both of which are essential for effective vision-language grounding. Compared to modular VLMs, where image features interact later with text, our approach enables text-image attention in every layer of the model, retaining important visual information~\cite{bolya2025perception}. %
We fine-tune our early-fusion MoE VLM for referring expression segmentation (RES) and detection (RED). For RES, we attach a lightweight segmentation head over text-conditioned visual features and optimize the Dice and Focal losses. For RED, a detection head regresses bounding box coordinates and size. Text modeling uses next-token prediction on the text expressions and special task tokens.

Initializing vision experts from distillation weights aids in solving downstream tasks with limited annotated data. However, we qualitatively observe degradation of dense feature maps. In DINOv3, this correlates with learning global representations (classification improves while segmentation degrades) as patch features align more with the CLS token. We hypothesize this is due to vision-text modeling and could harm grounding tasks. To preserve dense feature quality during adaptation, we use Gram anchoring against the frozen AMoE from distillation. Let $b$ index images in a batch, $N_b$ be the number of patches, $\hat F_b\in\mathbb{R}^{N_b\times C}$ and $\hat T_b\in\mathbb{R}^{N_b\times C}$ be row-wise $\ell_2$-normalized student and teacher patch features. Denoting the  Gram matrices $K^{S}_b=\hat F_b\hat F_b^\top$ and $K^{T}_b=\hat T_b\hat T_b^\top$, the anchoring loss is

\begin{equation}
\label{eq:gram-anchor}
\mathcal{L}_{\mathrm{gram}}
=
\frac{1}{B}\sum_{b=1}^{B}\;
\frac{1}{N_b^2}\,\bigl\lVert K^{S}_b - K^{T}_b \bigr\rVert_F^2.
\end{equation}

\begin{table*}[t]
\centering
\small
\setlength{\tabcolsep}{4pt}
\adjustbox{width=\textwidth}{
\begin{tabular}{l c l| c c c c c c c c | c c c c c c}
\toprule
\multicolumn{3}{c|}{\bf Method} &
\multicolumn{8}{c|}{\bf Image–Text Classification @ $512{\times}512$ (Top-1)} &
\multicolumn{6}{c}{\bf kNN Classification @ $512{\times}512$ (Top-1)} \\
\cmidrule(lr){1-3}\cmidrule(lr){4-11}\cmidrule(lr){12-17}
Model & Budget & Head &
IN & C101 & CUB & Food & Flow & DTD & Air & Avg &
IN & CUB & Food & DTD & Air & Avg \\
\midrule
\multirow{2}{*}{Teachers (ref.)} & \multirow{2}{*}{-}
& DINOv3   & \demph{79.73} & \demph{89.02} & \demph{85.03} & \demph{94.97} & \demph{86.89} & \demph{65.50} & \demph{81.83} & \demph{83.28}
                               & \demph{85.70} & \demph{90.24} & \demph{94.75} & \demph{80.48} & \demph{84.32} & \demph{87.10} \\
& & SigLIP2  & \demph{80.82} & \demph{87.92} & \demph{78.49} & \demph{96.67} & \demph{89.39} & \demph{69.66} & \demph{79.59} & \demph{83.22}
                               & \demph{86.20} & \demph{85.14} & \demph{96.86} & \demph{77.34} & \demph{92.06} & \demph{87.52} \\
\midrule
\multirow{3}{*}{\parbox[c]{2cm}{RADIOv2.5-H \\ (0.6B)}} & \multirow{3}{*}{1.1TT}
 & CLIP     & 78.69 & 88.69 & 81.47 & 94.09 & 88.23 & 69.57 & 70.32 & 81.58
                               & 84.44 & 83.90 & 94.95 & 78.35 & 77.70 & 83.87 \\
& & SigLIP   & 77.03 & 88.42 & 73.00 & 94.02 & 87.26 & 70.20 & 64.18 & 79.16
                               & 84.14 & 82.79 & 95.24 & 76.97 & 74.82 & 82.79 \\
& & Ensemble & 78.94 & 88.53 & 81.78 & 94.31 & \textbf{89.32} & \textbf{72.68} & 70.26 & 82.26
                               & 85.71 & 86.07 & 95.30 & 78.88 & 79.62 & 85.12 \\
\midrule
TIPS (ViT-g, 1.1B) & 0.46TT & Ensemble & 72.47 & 87.14 & 75.09 & 89.57 & 79.25 & 57.13 & 16.13 & 68.11
                               & 83.50 & 87.40 & 93.90 & 76.30 & 66.60 & 84.50 \\
\midrule
\textbf{SigLino-0.6B} & 0.23TT & Ensemble & \textbf{80.50} & \textbf{89.90} & 83.00 & 95.00 & 89.00 & 71.10 & 82.40 & \textbf{84.40}
                               & 86.10 & 89.20 & 95.80 & 81.10 & 92.30 & \textbf{90.70} \\
\midrule
\textbf{SigLino-MoE-0.15-0.6B} & 0.23TT & Ensemble & 78.80 & 88.60 & 80.90 & 93.40 & 89.00 & 69.50 & 81.40 & \textit{83.10}
                               & 85.00 & 88.20 & 94.60 & 79.90 & 91.50 & \textit{89.80} \\
\midrule
\multirow{3}{*}{\parbox[c]{2cm}{\textbf{SigLino-MoE-0.3-0.6B}}}& \multirow{3}{*}{0.23TT}
 & DINOv3   & 78.26 & 87.71 & 82.86 & 93.36 & 87.50 & 65.00 & 76.86 & 81.65
                               & 84.21 & 89.26 & 94.15 & 80.11 & 81.65 & 85.88 \\
& & SigLIP2  & 77.66 & 88.02 & 74.05 & 94.06 & 89.56 & 67.66 & 76.59 & 81.09
                               & 84.14 & 81.79 & 95.29 & 75.85 & 90.05 & 85.42 \\
& & Ensemble & 80.17 & 88.76 & \textbf{82.78} & \textbf{94.67} & 89.20 & 70.16 & \textbf{83.18} & \textit{84.13}
                               & \textbf{85.89} & \textbf{88.83} & \textbf{95.56} & \textbf{79.26} & \textbf{90.77} & 88.06 \\
\bottomrule
\vspace{-0.5cm}
\end{tabular}
}
\caption{Per-benchmark classification at $512{\times}512$ comparing RADIOv2.5-H and our SigLino, with teacher references. We report per-dataset Top-1 and per-block macro-averages (Avg). We note that we outperform the teachers on average with the ensembled evaluations.\vspace{-0.4cm}}
\label{tab:cls_per_benchmark_512_twocol}
\end{table*}

\section{Experiments}
\label{sec:xp}

We evaluate image-text and $k$NN classification on ImageNet~\cite{russakovsky2015imagenet}, CUB-200~\cite{wah2011caltech}, Food-101~\cite{bossard2014food}, Flowers-102~\cite{nilsback2008automated}, DTD~\cite{cimpoi2014describing}, FGVC-Aircraft~\cite{maji2013fine}, and Caltech101~\cite{fei2004learning}; retrieval (Recall@1) on MSCOCO5k~\cite{lin2014microsoft} and Flickr30k~\cite{young2014image}; segmentation mIoU (linear probing at $512^2$) on ADE20k~\cite{zhou2017scene}, PASCAL-VOC~\cite{everingham2010pascal}, and Cityscapes~\cite{cordts2016cityscapes}; and grounding on RefCOCO/+/g~\cite{yu2016modeling,kazemzadeh2014referitgame}. We also evaluate our early-fusion Grounding VLM for segmentation and detection. In ablations, for ImageNet $k$NN evaluation, we use 100k training images subsampled from the original set.

\noindent \textbf{Teacher-heads ensembling:} For agglomerative models we use entropy-weighted ensembling: weights $\alpha_t(x)\propto \exp(-\gamma\,H_t(x))$ down-weight uncertain (high-entropy) heads, giving $\mathbf{s}_{\mathrm{ens}}(x)=\sum_{t}\alpha_t(x)\,\mathbf{s}_t(x)$. Full details are in the supplementary material. 

\noindent \textbf{Implementation details:}
We train on four nodes with eight A100 GPUs each, using sequence packing (up to 16 images per sequence) and a per-rank batch size of 24. Our SigLino student is an 18-layer MoE (0.3B active, 0.6B total) with 28 experts (6 active) and 768 dimensions, distilled in two stages: Stage 1 to 256×256 for 50k steps, Stage 2 to 768×768 for 90k steps. Alongside our main MoE model, we also train a dense counterpart (SigLino-Dense, 0.6B, replacing MoE layers with standard FFN) and an ultra-sparse MoE variant (SigLino-MoE-Sparse, top-2 of 28 experts, 0.15B active), all using the same training recipe, to study the efficiency–performance trade-off. 

\subsection{State-of-the-Art Comparison}

We compare SigLino against RADIOv2.5-L and H (0.3B and 0.6B parameters, respectively) baselines at comparable model scales, focusing on global representation quality at up to $512{\times}512$ pixels. We report per-dataset top-1 accuracy for image–text and $k$NN classification in Table~\ref{tab:cls_per_benchmark_512_twocol}, together with macro-averages. For retrieval, we report Recall@1 on MSCOCO5k and Flickr30k in Table~\ref{tab:ret_ms_f30k_512}. Teacher results (deemphasized) are for reference only.

\begin{table}[t]
\centering
\small
\resizebox{\columnwidth}{!}{
\begin{tabular}{l l cc}
\toprule
Method & Cityscapes $\uparrow$ & ADE20k $\uparrow$& PASCAL-VOC $\uparrow$\\
\midrule
{RADIOv2.5-L (0.3B)} & 62.47 & 50.95 & 84.83 \\
{RADIOv2.5-H (0.6B)} & 64.11 & 51.37 & 85.70 \\
TIPS (ViT-g, 1.1B) & 59.78 & 49.40 & 83.10 \\
\midrule
{SigLino-0.6B (Ours)} & \textbf{65.38} & \textbf{52.95} & \textbf{87.71} \\
{SigLino-MoE-0.15-0.6B (Ours)} & 63.94 & 51.28 & 86.76 \\
{SigLino-MoE-0.3-0.6B (Ours)} & \textit{64.36} & \textit{52.23} & \textit{87.05} \\
\bottomrule
\end{tabular}
\vspace{-0.4cm}
}
\caption{mIoU results on linear probing segmentation at $512^2$ resolution. Our dense and MoE models outperform baselines including TIPS~\cite{maninis2024tips}.\vspace{-0.2cm}}
\label{tab:linear_prob_seg}
\end{table}
\begin{table}[t]
\centering
\small
\setlength{\tabcolsep}{9pt}
\adjustbox{width=\columnwidth}{
\begin{tabular}{l l c c c c}
\toprule
\multirow{2}{*}{Method} & \multirow{2}{*}{Head} & \multicolumn{2}{c}{MSCOCO5k} & \multicolumn{2}{c}{Flickr30k} \\
\cmidrule(lr){3-4}\cmidrule(lr){5-6}
& & T2I $\uparrow$ & I2T $\uparrow$ & T2I $\uparrow$ & I2T $\uparrow$ \\
\midrule
\multirow{2}{*}{Teachers (ref.)} & DINOv3 & \demph{47.66} & \demph{64.44} & \demph{76.70} & \demph{90.70} \\
                                & SigLIP2 & \demph{52.10} & \demph{67.42} & \demph{78.58} & \demph{92.60} \\
                                 
\midrule
\multirow{3}{*}{\parbox[c]{2cm}{RADIOv2.5-L\\ (0.3B)}} & CLIP & 51.60 & 69.42 & 78.18 & 92.50 \\
                             & SigLIP & 49.94 & 67.60 & 77.76 & 92.20 \\
                             & Ens. & 52.44 & 71.04 & 79.82 & 93.10 \\
\midrule
\multirow{3}{*}{\parbox[c]{2cm}{RADIOv2.5-H \\ (0.6B)}}& CLIP & 52.24 & 70.92 & 79.26 & 92.90 \\
                             & SigLIP & 50.88 & 67.66 & 79.08 & 92.20 \\
                             & Ens. & 53.24 & 71.82 & 80.96 & 93.50 \\
\midrule
\textbf{SigLino-0.6B} & Ens. & \textbf{55.60} & \textbf{72.90} & \textbf{81.90} & \textit{94.20} \\
\midrule
\textbf{SigLino-MoE-0.15-0.6B} & Ens. & \textit{54.20} & \textit{71.10} & \textit{81.00} & 92.90 \\
\midrule
\multirow{3}{*}{\parbox[c]{2cm}{\textbf{SigLino-MoE-0.3-0.6B}}} & DINOv3 & 46.67 & 65.80 & 76.44 & 91.00 \\
                             & SigLIP2 & 51.81 & 68.18 & 78.58 & 91.90 \\
                             & Ens. & 53.98 & 72.14 & 81.20 & \textbf{94.30} \\
\bottomrule
\end{tabular}
\vspace{-0.3cm}}
\caption{Retrieval at $512{\times}512$ on MSCOCO5k and Flickr30k (Recall@1). Teacher rows are reference baselines.\vspace{-0.2cm}}
\label{tab:ret_ms_f30k_512}
\end{table}

\begin{table*}[t]
\centering
\small
\resizebox{\textwidth}{!}{
\begin{tabular}{l l c c c c c c}
\toprule
Method & Head & Img-Text Avg $\uparrow$ & kNN Avg $\uparrow$ & MSCOCO5k T2I@1 $\uparrow$ & MSCOCO5k I2T@1 $\uparrow$ & Flickr30k T2I@1 $\uparrow$ & Flickr30k I2T@1 $\uparrow$ \\
\midrule
\multirow{2}{*}{Vanilla MT} 
& DINOv3   & 63.71 & 81.57 & 38.78 & 53.76 & 66.22 & 82.30 \\
\multirow{2}{*}{(No RKD)}& SigLIP2  & 76.72 & 80.40 & 45.69 & 61.12 & 71.00 & 84.80 \\
& Ensemble & 77.62 & 83.54 & 48.15 & 64.10 & 74.30 & 89.10 \\
\midrule
\multirow{2}{*}{RKD} 
& DINOv3   & 77.48 & 81.36 & 42.17 & 60.16 & 70.22 & 85.80 \\
\multirow{2}{*}{(Symmetric)}& SigLIP2  & 76.05 & 79.61 & 45.31 & 60.26 & 70.12 & 84.30 \\
& Ensemble & 79.49 & 82.61 & 48.32 & \textbf{66.28} & 74.70 & \textbf{89.50} \\
\midrule
\multirow{2}{*}{ARKD} 
& DINOv3   & 77.68 & 81.99 & 42.68 & 60.52 & 69.86 & 86.70 \\
\multirow{2}{*}{(Asymmetric)}& SigLIP2  & 76.62 & 80.44 & 45.11 & 59.82 & 71.36 & 83.60 \\
 & Ensemble & \textbf{80.21} & \textbf{83.63} & \textbf{48.51} & 65.92 & \textbf{74.90} & 89.40 \\
\bottomrule
\vspace{-0.5cm}
\end{tabular}
}
\caption{ARKD improves image–text alignment over no RKD (largest gains for DINOv3) while preserving kNN clustering quality; vanilla RKD degrades kNN.\vspace{-0.4cm}}
\label{tab:rkd_ablation_256}
\end{table*}

\begin{table}[t]
\centering
\small
\resizebox{\columnwidth}{!}{
\begin{tabular}{l ccc ccc}
\toprule
& \multicolumn{3}{c}{Detection (Acc@IoU0.5) $\uparrow$} & \multicolumn{3}{c}{Segmentation (Acc@IoU0.5) $\uparrow$} \\
\cmidrule(lr){2-4}\cmidrule(lr){5-7}
Method & Ref & Refg & Ref+ & Ref & Refg & Ref+ \\
\midrule
Scratch       & 29.15 & 21.80 & 17.76 & 23.64 & 15.31 & 13.45 \\
SigLino init       & \textit{57.49} & \textit{45.58} & \textit{41.55} & \textit{57.74} &\textit{45.04} & \textit{39.81} \\
SigLino init +Gram  & \textbf{61.06} & \textbf{48.77} & \textbf{47.09} & \textbf{63.38} & \textbf{50.37} & \textbf{46.48} \\
\bottomrule
\end{tabular}
\vspace{-0.5cm}
}
\caption{Referring expression grounding results. Distillation substantially improves over scratch training; adding Gram anchoring furthers gains across RefCOCO, RefCOCOg, and RefCOCO+.\vspace{-0.35cm}}
\label{tab:grounding_results}
\end{table}

\noindent \textbf{Multi-teacher distillation for grounding:} Table~\ref{tab:multiteacher_grounding} shows that multi-teacher initialization substantially outperforms single-teacher: SigLIP2-only yields 40.69 RefCOCO, DINOv3-only 45.06, while SigLino reaches 54.72 ($+9.66$ over DINOv3-only), demonstrating the complementarity of vision-text alignment and dense features for grounding.

\begin{table}[t]
\centering
\small
\resizebox{\columnwidth}{!}{
\begin{tabular}{l|ccc}
\toprule
Student Teacher(s) & RefCOCO & RefCOCOg & RefCOCO+ \\
\midrule
SigLIP2 only & 40.69 & 28.26 & 24.29 \\
DINOv3 only & \textit{45.06} & \textit{31.27} & \textit{27.18} \\
\textbf{Multi-Teacher (SigLino)} & \textbf{54.72} & \textbf{37.23} & \textbf{33.01} \\
\bottomrule
\end{tabular}
}
\vspace{-0.7em}
\caption{Impact of multi-teacher vs. single-teacher distillation on referring expression segmentation (RES). Multi-teacher initialization substantially improves grounding performance.\vspace{-0.2cm}}
\label{tab:multiteacher_grounding}
\end{table}

\noindent \textbf{Overall comparison:}
Against RADIOv2.5 at comparable model scales, our SigLino-MoE sets a new state-of-the-art, surpassing RADIOv2.5-H on macro-averaged image–text classification (84.13 vs.\ 82.26) and kNN (88.06 vs.\ 85.12) while using $\sim$230 billion image tokens—4.7$\times$ fewer than RADIO's 1.1 trillion. Our dense variant (SigLino-Dense) further improves to 84.40 image-text and 90.70 kNN, confirming the quality of our distillation pipeline, while SigLino-MoE with 6 active experts achieves very close performance at only half the active parameters—representing the best efficiency–performance trade-off in our model family. The ultra-sparse MoE (top-2/28, 0.15B active) still outperforms RADIOv2.5-H (83.10 / 89.80). On fine-grained classification, SigLino excels on FGVC-Aircraft; on retrieval, it outperforms on MSCOCO5k and Flickr30k; all three variants further improve on linear probing segmentation (see Tables~\ref{tab:cls_per_benchmark_512_twocol}, \ref{tab:ret_ms_f30k_512}, \ref{tab:linear_prob_seg}).  

\noindent \textbf{Ensembling:}
Our per-head results are more balanced than RADIO’s, and the ensembling consistently yields larger gains, indicating stronger head complementarity. At $512^2$, SigLino improves substantially over each head on both image–text and $k$NN (Table~\ref{tab:cls_per_benchmark_512_twocol}), and exceeds teacher references on macro-averages and on retrieval (Table~\ref{tab:ret_ms_f30k_512}). This is consistent with the intended effect of relation-aware distillation.

\noindent \textbf{Referring expression:}
Table \ref{tab:grounding_results} shows that SigLino initialization delivers large gains over scratch training, roughly doubling to tripling performance across all referring benchmarks in both detection and segmentation. Table \ref{tab:multiteacher_grounding} demonstrates that multi-teacher initialization substantially outperforms single-teacher distillation. 

\subsection{Ablations}

\noindent \textbf{Impact of ARKD:} Table~\ref{tab:rkd_ablation_256} confirms that relational KD consistently boosts image–text alignment, with the largest gains for DINOv3. Vanilla RKD slightly degrades $k$NN; ARKD recovers clustering quality while preserving alignment gains, yielding the best overall trade-off. The marginal SigLIP2 drop is attributed to ARKD rebalancing student capacity across teachers.

\noindent \textbf{Impact of OpenLVD200M:} We ablate our data curation pipeline by comparing OpenLVD200M against a random uniform subsample of equal size and reporting results in Table~\ref{tab:openlvd_main}. In image–text classification, the curated set yields consistent gains: the average accuracy rises from 74.96 to 79.11 (+4.15), with significant improvements on fine-grained/long-tail datasets (FGVC-Aircraft, +18.64; see supplementary for per-benchmark breakdown).

\begin{table}[t]
\centering
\small
\setlength{\tabcolsep}{5pt}
\adjustbox{width=0.8\columnwidth}{
\begin{tabular}{lccccc}
\toprule
Method & IT $\uparrow$ & kNN $\uparrow$ & T2I $\uparrow$ & I2T $\uparrow$  \\
\midrule
Random (200M) & 74.96 & 82.66 & 57.63 & 75.12 \\
OpenLVD200M & \textbf{79.11} & \textbf{85.08} & \textbf{59.14} & \textbf{76.43} \\
\bottomrule
\vspace{-0.4cm}
\end{tabular}
}
\caption{Curated \vs random sampling (ensemble student). Reported results are macro-averages across benchmarks. \vspace{-0.4cm}}
\label{tab:openlvd_main}
\end{table}

\noindent \textbf{Impact of Gram-anchoring:}
Gram‑anchoring yields consistent gains over plain fine‑tuning (Table~\ref{tab:grounding_results}). PCA maps (Figure~\ref{fig:gram-anchoring}) show that naive fine-tuning blurs the dense structure while Gram-anchoring restores spatial coherence, consistent with our premise that preserving inter-patch geometry improves grounding.

\vspace{-0.5em}
\section{Conclusion}
\vspace{-0.5em}
\label{sec:ccls}
We present SigLino, a family of vision foundation models built on efficient multi‑teacher distillation from SigLIP2 and DINOv3, featuring data curation, asymmetric relational knowledge distillation, and token‑balanced batching. Our SigLino models achieve improved performances over existing agglomerative models on classification, image-text matching and segmentation tasks. Furthermore, initializing the vision experts of a grounding early-fusion MoE VLM achieves improved referring expression detection and segmentation results over training from scratch.

{
    \small
    \bibliographystyle{ieeenat_fullname}
    \bibliography{main}
}

\clearpage
\setcounter{page}{1}
\maketitlesupplementary

\section{Supplementary Results: Small Model Sizes}
\label{sec:intermediate_models}

To explore the trade-off between model size and performance, we evaluate two intermediate-scale models distilled using the same recipe as the main SigLino-0.6B: SigLino-30m (30M parameters) and SigLino-70m (70M parameters). These models validate the scalability of our distillation approach across different parameter budgets and provide useful baselines for compute-constrained settings. Results show that even at significantly reduced model sizes, the multi-teacher distillation recipe maintains strong performance on classification, retrieval, and segmentation tasks (Tables \ref{tab:intermediate_cls}, \ref{tab:intermediate_ret}, and \ref{tab:intermediate_seg}).

\begin{table*}[t]
\centering
\small
\setlength{\tabcolsep}{5pt}
\begin{tabular}{l l | c c c c c c c | c c c c c c}
\toprule
\multicolumn{2}{c|}{\bf Model} & \multicolumn{7}{c|}{\bf Image–Text Classification @ $512{\times}512$ (Top-1)} & \multicolumn{6}{c}{\bf kNN Classification @ $512{\times}512$ (Top-1)} \\
\cmidrule(lr){1-2}\cmidrule(lr){3-9}\cmidrule(lr){10-15}
Model & Head &
IN & C101 &  Food & Flowers & DTD & Air & Avg &
IN & Food & DTD & Air & Avg \\
\midrule
\multirow{2}{*}{SigLino-30m}
& DINOv3   & 60.41 & 85.24 & 75.22 & 74.38 & 58.98 & 40.76 & 69.20
                               & 76.47 & 85.24 & 71.84 & 70.35 & 76.23 \\
& SigLIP2  & 64.22 & 86.54  & 78.85 & 74.88 & 63.09 & 48.26 & 73.76
                               & 78.47 & 87.49 & 75.74 & 77.28 & 80.23 \\
& Ensemble & 65.14 & 87.16  & 80.34 & 77.43 & 62.85 & 48.26 & 70.20
                               & 78.97 & 87.49 & 75.74 & 77.28 & 83.73 \\
\midrule
\multirow{2}{*}{SigLino-70m}
& DINOv3   & 67.91 & 87.22 & 82.42 & 81.34 & 63.81 & 56.44 & 75.50
                               & 80.12 & 89.66 & 77.34 & 82.01 & 82.28 \\
& SigLIP2  & 71.67 & 88.66 & 84.98 & 84.15 & 65.66 & 60.10 & 78.38
                               & 81.73 & 90.18 & 77.34 & 83.30 & 83.14 \\
& Ensemble & 71.20 & 88.33 & 85.19 & 83.98 & 65.66 & 60.10 & 75.74
                               & 81.73 & 90.18 & 77.34 & 83.30 & 86.44 \\
\bottomrule
\end{tabular}
\caption{Classification performance for small model sizes at $512{\times}512$ resolution. SigLino-30m and SigLino-70m demonstrate the effectiveness of our distillation recipe across different parameter budgets.}
\label{tab:intermediate_cls}
\end{table*}

\begin{table}[t]
\centering
\small
\setlength{\tabcolsep}{9pt}
\resizebox{\columnwidth}{!}{
\begin{tabular}{l l c c c c}
\toprule
\multirow{2}{*}{Model} & \multirow{2}{*}{Head} & \multicolumn{2}{c}{MSCOCO5k} & \multicolumn{2}{c}{Flickr30k} \\
\cmidrule(lr){3-4}\cmidrule(lr){5-6}
& & T2I@1 & I2T@1 & T2I@1 & I2T@1 \\
\midrule
\multirow{3}{*}{SigLino-30m}
& DINOv3   & 38.90 & 49.18 & 65.66 & 78.40 \\
& SigLIP2  & 43.38 & 56.22 & 71.18 & 82.80 \\
& Ensemble & 46.58 & 59.72 & 72.86 & 82.20 \\
\midrule
\multirow{3}{*}{SigLino-70m}
& DINOv3   & 43.66 & 61.10 & 70.88 & 85.80 \\
& SigLIP2  & 47.01 & 58.30 & 77.32 & 90.20 \\
& Ensemble & 50.42 & 65.44 & 77.54 & 90.50 \\
\bottomrule
\end{tabular}
}
\caption{Retrieval performance (Recall@1) for intermediate model sizes on MSCOCO5k and Flickr30k at $512{\times}512$ resolution.}
\label{tab:intermediate_ret}
\end{table}

\begin{table}[t]
\centering
\small
\resizebox{\columnwidth}{!}{
\begin{tabular}{l c c c}
\toprule
Model & ADE20k $\uparrow$ & PASCAL-VOC $\uparrow$ & Parameters \\
\midrule
SigLino-30m & 43.68 & 82.43 & 30M \\
SigLino-70m & 45.33 & 84.26 & 70M \\
\bottomrule
\end{tabular}
}
\caption{Segmentation performance (mIoU) for intermediate model sizes on linear probing at $512^2$ resolution. These smaller models demonstrate that the distillation recipe scales effectively across different parameter budgets.}
\label{tab:intermediate_seg}
\end{table}

\section{Analysis of PHI-S Transformation on Registers}
\label{sec:phi-s-analysis}
\begin{figure*}[t]
    \centering
    \includegraphics[width=\textwidth]{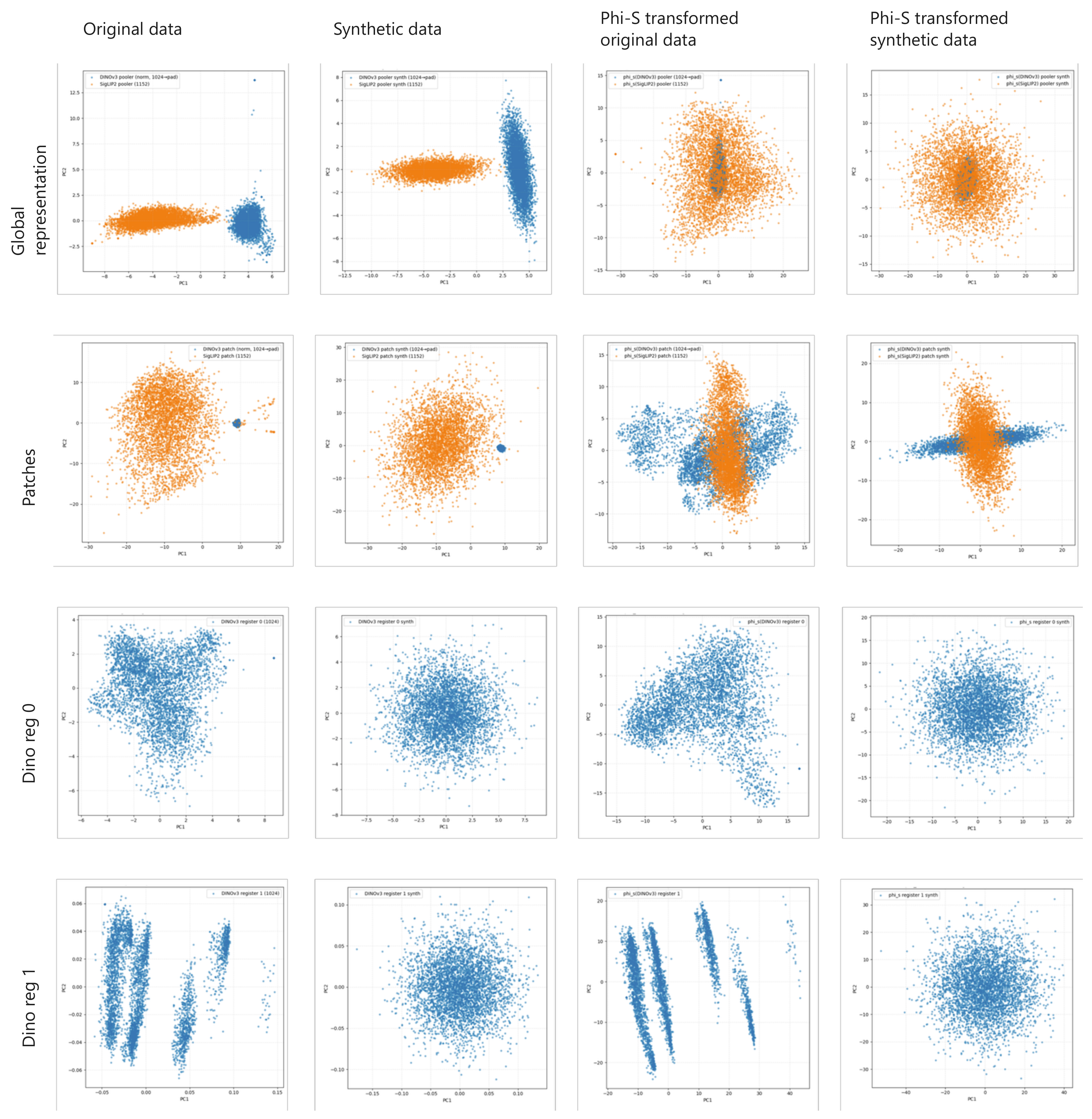}
    \caption{We visualize PCA projections of global features, patches, and DINOv3 registers (0 and 1): original data (Col 1), synthetic Gaussian data generated from estimated moments (Col 2), and their respective versions after Phi-S transformation (Cols 3 and 4). While global, patch embeddings, and the 0th register are well-approximated by Gaussian statistics and effectively whitened by Phi-S, the first register exhibits multi-mode distributions (Row 4) where simple moments capture inter-mode statistics. Hence, applying Phi-S to this register yields incorrect transformations.}
    \label{fig:phis_pca}
\end{figure*}

We apply PHI-S~\cite{ranzinger2024phi} to evenly distribute the statistical influence of diverse channels and teacher representations. PHI-S operates by rotating the feature space via an invertible transform, composed of PCA whitening and a Hadamard rotation, such that the variance is distributed uniformly across all channels. This normalization assumes that the underlying feature distributions can be reasonably approximated by their first and second-order moments (\textit{i.e.}, Gaussian-like). While this assumption holds for global summary tokens and patch embeddings, we observe that the DINOv3 first register token has a multi-mode distribution. As illustrated in Figure~\ref{fig:phis_pca}, the first register (Row 4) forms distinct, separated clusters. Thus, standard moment estimation captures the statistics between these modes rather than the variance within them. This discrepancy is highlighted by the synthetic data generated from these estimated moments (Column 2), which fails to reproduce the structure of the original data (Column 1) as compared to the zeroth register, global, and patch representations. When PHI-S is applied based on these ill-fitted statistics, it results in a transformed distribution (Column 3) that diverges significantly from the intended standardized target (Column 4). In practice, forcing this transformation on this multi-mode register leads to incorrect scaling and centering, resulting in training instability. Therefore, we exclude registers from the PHI-S normalization pipeline and supervise them in their original space. 

\section{Impact of Asymmetric Relational Knowledge Distillation (ARKD)}
\label{sec:arkd_analysis}
\begin{figure*}[t]
    \centering
    \includegraphics[width=\textwidth]{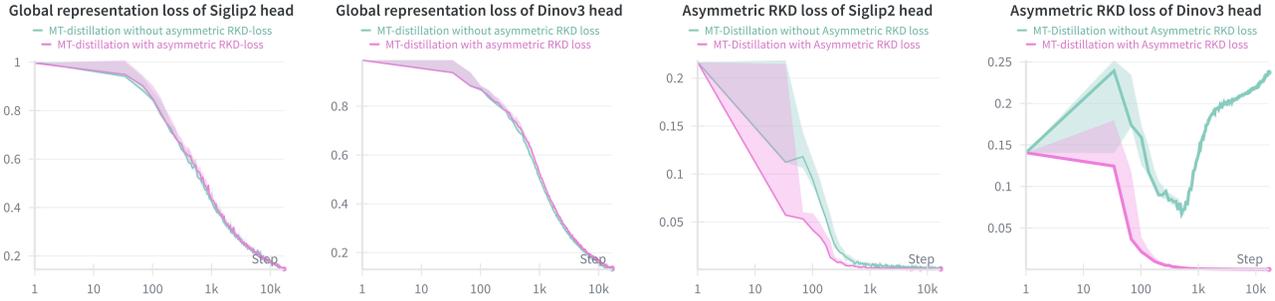}
    \caption{Impact of Asymmetric Relational Knowledge Distillation (ARKD) on training dynamics.}
    \label{fig:arkd-loss}
\end{figure*}
As introduced in the main text (Section~\ref{sec:mt-loss}), we propose Asymmetric Relational Knowledge Distillation (ARKD) to enforce pairwise geometric consistency in the student embedding space. Here, we provide an empirical analysis of its effect on training dynamics. Figure~\ref{fig:arkd-loss} visualizes the evolution of both global representation (cosine) losses and relational (ARKD) losses throughout training, comparing a model trained with the full AMoE objective (pink) against a baseline trained without the ARKD term (green).

For SigLIP2 (plots 1 and 3), the global loss and relational loss decrease together even without explicit relational supervision, suggesting that SigLIP2's contrastive objective naturally induces a consistent pairwise structure. However, for DINOv3 (plots 2 and 4), in the baseline experiment (green curve, rightmost plot), the relational error actually fluctuates in both directions as the global cosine loss is optimized. This indicates that DINOv3's pointwise supervision alone is insufficient to preserve the teacher’s geometry. 

By explicitly optimizing the ARKD objective (pink curve), we force the student to respect these pairwise constraints. The loss trajectory shows that ARKD acts as a regularizer, enforcing relational geometry between samples. This enforced structural alignment directly correlates with the significant improvements observed in zero-shot image-text classification for the DINOv3 head.

\section{Positional Encoding Analysis}
\label{sec:rope_analysis}

\begin{figure*}[t]
    \centering
    \includegraphics[width=0.85\textwidth, keepaspectratio]{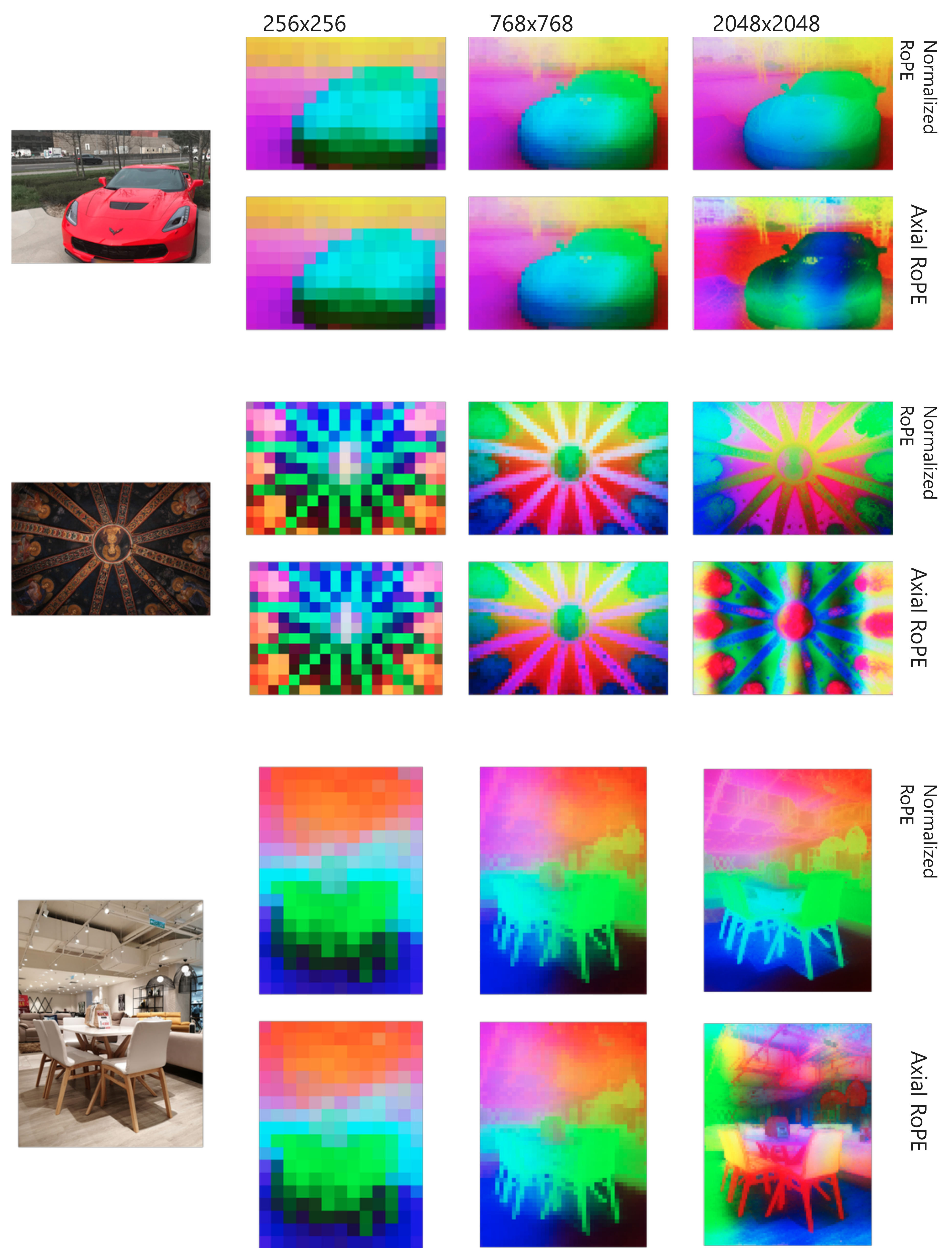}
    \caption{Impact of positional encoding on unseen resolutions. We compare feature map consistency across resolutions ($256{\times}256$ to $2048{\times}2048$ pixels) for Normalized RoPE (top) versus standard Axial RoPE (bottom) using the distilled DINOv3 head. While both methods perform comparably at the training resolutions (up to $768{\times}768$ pixels), Axial RoPE degrades at high resolutions, losing object consistency and introducing artifacts. In contrast, Golden RoPE maintains strong scale invariance and feature coherence even at extreme, unseen resolutions ($2048{\times}2048$ pixels, \textit{i.e.}, 16k patches), demonstrating better extrapolation capabilities for MT-distillation.}
    \label{fig:golden_rope}
\end{figure*}

We investigate the impact of the Rotary Positional Embedding (RoPE) strategy on the student's ability to generalize to unseen high resolutions. Specifically, we compare the standard Axial RoPE against normalizing the input coordinates based on the image aspect ratio (mapping coordinates roughly to $[-1, 1]$) rather than using absolute integer indices. This ensures that the relative frequency distribution remains consistent regardless of the absolute image resolution.
Figure~\ref{fig:golden_rope} demonstrates the generalization capabilities of both methods. We visualize the feature maps of the distilled DINOv3 head across resolutions ranging from the training size ($256\times256$) to an unseen high resolution ($2048\times2048$). 
With standard Axial RoPE (bottom row), we observe a breakdown in feature coherence at high resolutions: the global structure degrades, and grid-like artifacts appear; the model struggles to extrapolate the axis-aligned frequencies beyond the training distribution.
In contrast, the normalized version (top row) exhibits strong scale invariance and good generalization on unseen resolutions. The feature maps at $2048\times2048$ retain the semantics and smoothness of the low-resolution inputs. 

\begin{figure}[t]
    \centering
    \includegraphics[width=0.9\columnwidth]{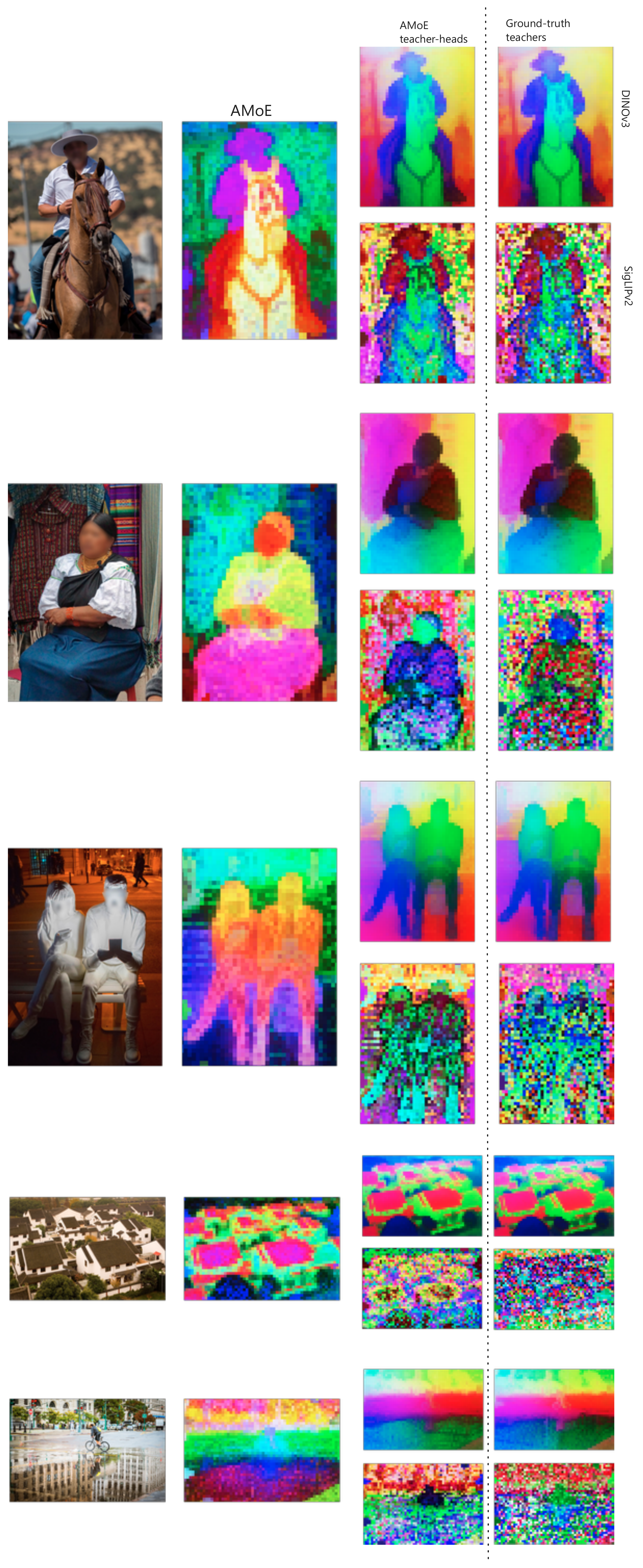}
    \caption{PCA-maps of learned representations: the original image, the shared SigLino backbone features, the student's teacher-specific projections (top: DINOv3 head, bottom: SigLIP2 head), and the corresponding ground-truth teacher features. The student closely reconstructs the teacher's distributions.}
    \label{fig:teacher-gt}
\end{figure}

\section{Expert Specialization Analysis via Linear CKA}

To investigate the semantic specialization of individual experts within the student model, we analyze the similarity between the representations routed to each expert and the hierarchical features of our teacher models (e.g., SigLIP2, DINOv3). We use \textbf{Linear Centered Kernel Alignment (CKA)}~\cite{kornblith2019similarity} as our similarity metric, chosen for its invariance to orthogonal transformations and isotropic scaling, making it suitable for comparing representation spaces of differing dimensions.

\paragraph{Experimental Protocol.}
For a given MoE layer in the student model, we iterate through 1k images. For each expert $e$, we aggregate the set of token embeddings $\mathbf{X}_e$ that the router assigns to that expert. Simultaneously, we extract the spatially corresponding token embeddings $\mathbf{Y}_{e, l}$ from layer $l$ of the teacher model. This spatial alignment ensures that we compare the student's routed features directly against the teacher's representation of the exact same image patches.

\paragraph{Formulation.}
Linear CKA measures the similarity between these two sets of representations based on the Frobenius norm of their cross-covariance matrix. Formally, for the collection of $N$ tokens routed to expert $e$ across the entire dataset, we compute:

\begin{equation}
    \text{CKA}(\mathbf{X}_e, \mathbf{Y}_{e, l}) = \frac{\| \text{cov}(\mathbf{X}_e, \mathbf{Y}_{e, l}) \|_F^2}{\| \text{cov}(\mathbf{X}_e, \mathbf{X}_e) \|_F \| \text{cov}(\mathbf{Y}_{e, l}, \mathbf{Y}_{e, l}) \|_F}
\end{equation}

where $\|\cdot\|_F$ denotes the Frobenius norm, and the centered cross-covariance matrix is defined as $\text{cov}(\mathbf{A}, \mathbf{B}) = \mathbf{A}^\top \mathbf{B} - \frac{1}{N}(\sum \mathbf{a}_i)(\sum \mathbf{b}_i)^\top$.

\subsection{Analysis of Expert Specialization}

\begin{figure}[t]
    \centering
    \includegraphics[width=\columnwidth, keepaspectratio]{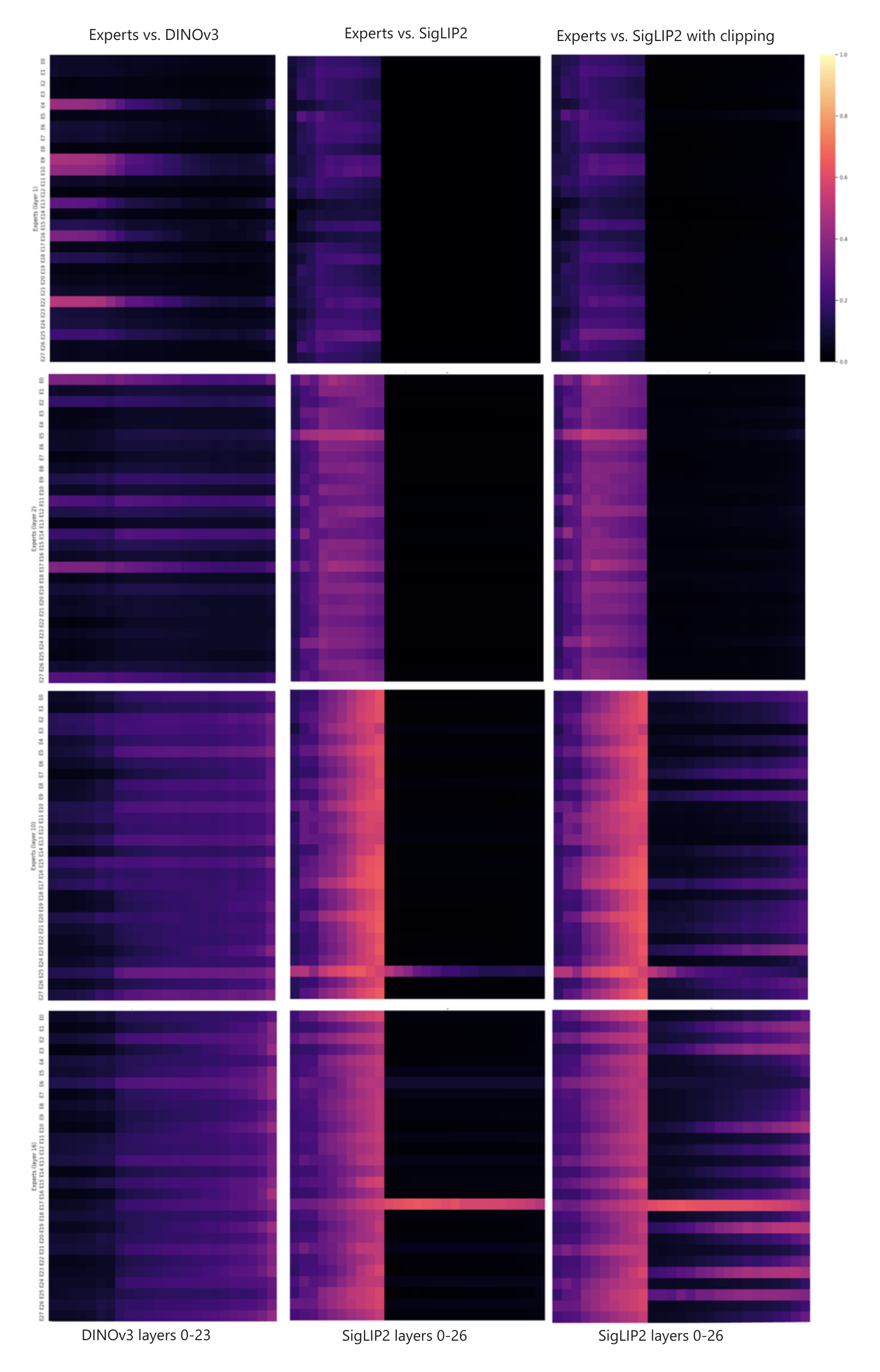}
    \caption{Linear CKA alignments between MoE experts and teacher layers at several SigLino layers.}
    \label{fig:linear_cka}
\end{figure}

Figure~\ref{fig:linear_cka} visualizes the Linear CKA alignment between the routed inputs of MoE experts at various depths (layers 1, 2, 10, 16) and the hierarchical representations of our teacher models, SigLIP2 and DINOv3. 
First, we observe a clear layer-wise progression: earlier student layers (e.g., Layers 1 and 2) align primarily with the shallow layers of the teachers, while deeper student layers shift their alignment towards the final teacher representations. This trend is particularly pronounced for SigLIP2, where student experts in early layers focus entirely on the first $\approx 10$ teacher layers. This is likely due to the emergence of high-magnitude activations in SigLIP2's deeper layers (potentially from the absence of register tokens).

More importantly, our analysis reveals teacher-specific specialization among experts, validating the choice of the Mixture-of-Experts architecture for multi-teacher distillation. 
In early layers, certain experts specialize exclusively in one teacher's features. For instance, in Layer 1, experts E4 and E22 show strong alignment with DINOv3 but low correlation with SigLIP2, whereas E5 specializes in SigLIP2 features. Similarly, in Layer 2, E5 is highly aligned with SigLIP2 while showing low similarity to DINOv3. We also observe shared experts that maintain alignment with both feature spaces.

In deeper layers (Layers 10 and 16), the specialization mechanism adapts to handle the high-magnitude activations characteristic of the SigLIP2 teacher. We observe a subset of experts, such as E25 in Layer 10 and E17 in Layer 16, that are strongly aligned with the latest layers of SigLIP2. These experts seem to be responsible for injecting these high-norm features into the student's representation space. 
Interestingly, other experts in these deep layers initially appear unaligned with SigLIP2. However, when we clip the teacher representations to the range $[-10, 10]$ (third column), we observe some alignments (e.g., experts E25 and E26 in Layer 16). This indicates that while a few experts handle the extreme value distribution, others continue to process the underlying semantic content of the SigLIP2 features, confirming that teacher-specific specialization persists throughout the network depth.

\section{Qualitative Analysis of Distilled Representations}
\label{sec:qualitative_vis}

\begin{lstlisting}[language=Python, float=t, caption={SigLino forward pseudo-code}]
# 1. Student Architecture (Agglomerative-MoE)
def StudentForward(packed_tokens, packing_mask):
    # Input: Packed sequence of multiple images (Token-Balanced Batching)
    # 1. Prepend CLS + 4 Registers (DINOv3 style) per image in sequence
    x = AddSpecialTokens(packed_tokens, num_regs=4)
    
    # 2. MoE Backbone with FlexAttention (prevents inter-image attn)
    h_latent = MoETransformer(x, mask=packing_mask)

    # 3. Project features to Teacher Spaces via Learnable Adapters
    # DINOv3: Project all tokens (CLS, Regs, Patches) to 1024-dim
    z_dino = Adapter_DINO(h_latent)

    # SigLIP2: Project to 1152-dim, then apply Frozen Attention Pooling
    # Pooler uses a learned probe query attending only to valid patches
    h_siglip = Adapter_SigLIP(h_latent)
    z_sig_summ = FrozenSigLIPPooler(h_siglip, query=Probe, mask=packing_mask)
    z_sig_patch = h_siglip[patches_only]

    return { "dino": z_dino, "siglip": (z_sig_summ, z_sig_patch) }
\end{lstlisting}
\label{lst:forward}

We provide a qualitative comparison of the distilled student features against the teacher baselines in Figure~\ref{fig:teacher-gt}. This qualitative analysis demonstrates that we successfully learn both teacher representations with high fidelity and that the SigLino patch representations constitute a synthesis of SigLIP2 and DINOv3. The shared SigLino backbone (Column 2) demonstrates nice synergies. While SigLIP2 features often suffer from artifacts harming performance on dense downstream tasks, and DINOv3 lacks inherent image-text alignment, the student's backbone converges on a representation that balances these characteristics. It retains the text-aware features in SigLIP2 with the geometric consistency provided by DINOv3. The resulting feature maps appear to have better object discriminability compared to each teacher individually.
\vspace{-0.5em}
\section{Training Implementation Details}
\label{sec:training_details}

We train our 18-layer MoE student model ($d{=}768$, 28 experts, top-$k{=}6$) on 4 nodes with 8$\times$A100 GPUs each. We use the AdamW optimizer with $\beta_1{=}0.9$, $\beta_2{=}0.999$, and $\epsilon{=}10^{-15}$. The learning rate follows a linear decay schedule from $10^{-3}$ to $10^{-4}$ after a 500-step warmup, with weight decay set to $0.02$. We summarize the pseudo-code of the distillation pipeline in Listings~\ref{lst:forward} and \ref{lst:loss}. The algorithm outlines the Agglomerative-MoE student forward pass, detailing how shared backbone features are projected into distinct DINOv3 and SigLIP2 embedding spaces via teacher-specific adapters and pooling mechanisms. It also formalizes the calculation of our multi-objective loss, explicitly showing how dense feature alignment is normalized by per-image token counts and combined with the global Asymmetric Relational Knowledge Distillation (ARKD) term to ensure structural consistency across the token-balanced batch.
\vspace{-0.5em}
\section{Detailed Ablation Benchmarks}
\label{sec:full_ablation_results}

We provide the full per-dataset results for our ablations. Table~\ref{tab:ablation_openlvd_cls} and Table~\ref{tab:ablation_openlvd_ret} detail the comparison between our curated OpenLVD200M dataset and random subsampling, highlighting the consistent gains across fine-grained classification and retrieval tasks. Similarly, Table~\ref{tab:ablation_arkd_cls} and Table~\ref{tab:ablation_arkd_ret} present the full breakdown of the ARKD ablation.

\begin{table*}[t]
\centering
\small
\setlength{\tabcolsep}{4.5pt}
\begin{tabular}{l l | c c c c c c c c | c c c c c c}
\toprule
\multicolumn{2}{c|}{\bf Method} & \multicolumn{8}{c|}{\bf Image–Text Classification @ $256{\times}256$ (Top-1)} & \multicolumn{6}{c}{\bf kNN Classification @ $256{\times}256$ (Top-1)} \\
\cmidrule(lr){1-2}\cmidrule(lr){3-10}\cmidrule(lr){11-16}
Model & Head &
IN & C101 & CUB & Food & Flow & DTD & Air & Avg &
IN & CUB & Food & DTD & Air & Avg \\
\midrule
\multirow{3}{*}{Random}
& DINOv3   & 68.97 & 87.76 & 68.67 & 87.94 & 83.37 & 62.57 & 47.42 & 72.39
                               & 76.45 & 82.21 & 90.98 & 77.61 & 66.73 & 82.27 \\
& SigLIP2  & 66.42 & 88.36 & 59.21 & 86.31 & 78.16 & 64.01 & 50.21 & 70.38
                               & 71.60 & 69.76 & 90.72 & 73.30 & 66.61 & 78.54 \\
& Ensemble & 70.51 & \textbf{89.47} & 70.40 & \textbf{88.84} & 85.32 & 67.02 & 53.18 & 74.96
                               & 76.18 & 81.93 & \textbf{91.53} & 76.70 & 69.96 & 82.66 \\
\midrule
\multirow{3}{*}{OpenLVD}
& DINOv3   & 72.45 & 87.57 & \textbf{74.38} & 87.69 & 87.14 & 63.10 & 62.56 & 76.41
                               & 77.89 & \textbf{84.12} & 90.94 & \textbf{78.51} & 74.64 & 84.31 \\
& SigLIP2  & 70.29 & 88.12 & 63.38 & 86.10 & 86.17 & 64.84 & 66.49 & 75.06
                               & 74.25 & 73.02 & 90.51 & 74.36 & 79.80 & 81.89 \\
& Ensemble & \textbf{73.74} & 89.44 & 73.95 & 88.53 & \textbf{88.71} & \textbf{67.55} & \textbf{71.82} & \textbf{79.11}
                               & \textbf{78.07} & 83.33 & 91.32 & 77.23 & \textbf{80.76} & \textbf{85.08} \\
\bottomrule
\end{tabular}
\caption{Ablation of data curation strategy (OpenLVD200M vs. Random Uniform Sampling) on Image-Text and kNN classification tasks at $256{\times}256$ resolution. OpenLVD yields consistent gains across all benchmarks, especially on fine-grained tasks like FGVC-Aircraft.}
\label{tab:ablation_openlvd_cls}
\end{table*}

\begin{table*}[t]
\centering
\small
\setlength{\tabcolsep}{4.5pt}
\begin{tabular}{l l | c c c c c c c c | c c c c c c}
\toprule
\multicolumn{2}{c|}{\bf Method} & \multicolumn{8}{c|}{\bf Image–Text Classification @ $256{\times}256$ (Top-1)} & \multicolumn{6}{c}{\bf kNN Classification @ $256{\times}256$ (Top-1)} \\
\cmidrule(lr){1-2}\cmidrule(lr){3-10}\cmidrule(lr){11-16}
Loss & Head &
IN & C101 & CUB & Food & Flow & DTD & Air & Avg &
IN & CUB & Food & DTD & Air & Avg \\
\midrule
\multirow{3}{*}{Vanilla}
& DINOv3   & 63.00 & 85.00 & 39.59 & 75.24 & 81.31 & 58.28 & 43.56 & 63.71
                               & 78.13 & 84.26 & 91.06 & 78.51 & 75.87 & 81.57 \\
& SigLIP2  & 71.03 & 87.92 & 66.81 & 85.64 & 87.38 & 64.88 & 73.41 & 76.72
                               & 74.97 & 76.16 & 90.74 & 74.36 & \textbf{85.79} & 80.40 \\
& Ensemble & 72.03 & 88.58 & 69.07 & 85.67 & 87.99 & 66.51 & 73.53 & 77.62
                               & \textbf{79.07} & 84.41 & \textbf{91.70} & 77.18 & 85.34 & 83.54 \\
\midrule
\multirow{3}{*}{RKD}
& DINOv3   & 72.57 & 87.86 & \textbf{76.64} & 87.58 & 87.14 & 63.67 & 66.94 & 77.48
                               & 77.71 & 84.33 & 90.87 & 77.77 & 76.11 & 81.36 \\
& SigLIP2  & 70.61 & 88.32 & 67.45 & 85.11 & 86.89 & 64.34 & 69.63 & 76.05
                               & 74.69 & 75.90 & 90.57 & 74.15 & 82.76 & 79.61 \\
& Ensemble & 74.07 & 89.15 & 75.84 & \textbf{88.03} & \textbf{88.96} & 66.70 & 73.65 & 79.49
                               & 78.10 & 84.21 & 91.42 & 76.81 & 82.52 & 82.61 \\
\midrule
\multirow{3}{*}{ARKD}
& DINOv3   & 72.75 & 88.29 & 75.93 & 87.66 & 86.89 & 63.81 & 68.44 & 77.68
                               & 78.05 & \textbf{84.91} & 91.04 & \textbf{79.10} & 76.83 & 81.99 \\
& SigLIP2  & 70.77 & 87.82 & 67.29 & 84.70 & 86.89 & 64.70 & 74.19 & 76.62
                               & 74.70 & 76.03 & 90.56 & 75.16 & 85.76 & 80.44 \\
& Ensemble & \textbf{74.28} & \textbf{89.24} & 76.17 & 87.97 & 88.71 & \textbf{67.45} & \textbf{77.67} & \textbf{80.21}
                               & 78.33 & 84.72 & 91.52 & 77.93 & 85.64 & \textbf{83.63} \\
\bottomrule
\end{tabular}
\caption{Ablation of Asymmetric vs. Symmetric Relational Knowledge Distillation (RKD) on classification tasks at $256{\times}256$. ARKD preserves the gains in image-text alignment from Symmetric RKD while recovering the kNN performance lost by the symmetric constraint.}
\label{tab:ablation_arkd_cls}
\end{table*}

\begin{table}[t]
\centering
\small
\setlength{\tabcolsep}{5pt}
\begin{tabular}{l lcccc}
\toprule
\multirow{2}{*}{Loss} & \multirow{2}{*}{Head} & \multicolumn{2}{c}{MSCOCO5k} & \multicolumn{2}{c}{Flickr30k} \\
\cmidrule(lr){3-4}\cmidrule(lr){5-6}
& & T2I@1 & I2T@1 & T2I@1 & I2T@1 \\
\midrule
\multirow{3}{*}{Vanilla}
& DINOv3   & 38.78 & 53.76 & 66.22 & 82.30 \\
& SigLIP2  & 45.69 & 61.12 & 71.00 & 84.80 \\
& Ensemble & 48.15 & 64.10 & 74.30 & 89.50 \\
\midrule
\multirow{3}{*}{Sym.\ RKD}
& DINOv3   & 42.17 & 60.16 & 70.22 & 85.80 \\
& SigLIP2  & 45.31 & 60.26 & 70.12 & 84.30 \\
& Ensemble & 48.32 & \textbf{66.28} & 74.70 & \textbf{89.50} \\
\midrule
\multirow{3}{*}{Asym.\ RKD}
& DINOv3   & 42.68 & 60.52 & 69.86 & 86.70 \\
& SigLIP2  & 45.11 & 59.82 & 71.36 & 83.60 \\
& Ensemble & \textbf{48.51} & 65.92 & \textbf{74.90} & 89.40 \\
\bottomrule
\end{tabular}
\caption{Impact of ARKD on retrieval (Recall@1) for MSCOCO5k and Flickr30k at $256{\times}256$. Relational distillation provides a significant boost over the Vanilla baseline, especially for the DINOv3 head.}
\label{tab:ablation_arkd_ret}
\end{table}

\begin{table}[t]
\centering
\small
\setlength{\tabcolsep}{5pt}
\begin{tabular}{l lcccc}
\toprule
\multirow{2}{*}{Method} & \multirow{2}{*}{Head} & \multicolumn{2}{c}{MSCOCO5k} & \multicolumn{2}{c}{Flickr30k} \\
\cmidrule(lr){3-4}\cmidrule(lr){5-6}
& & T2I@1 & I2T@1 & T2I@1 & I2T@1 \\
\midrule
\multirow{3}{*}{Random}
& DINOv3   & 42.87 & 60.22 & 69.94 & 87.00 \\
& SigLIP2  & 46.02 & 58.98 & 71.72 & 84.00 \\
& Ensemble & 48.78 & 65.86 & 74.58 & 89.80 \\
\midrule
\multirow{3}{*}{OpenLVD}
& DINOv3   & 43.62 & 60.94 & 72.32 & 88.70 \\
& SigLIP2  & 47.03 & 60.34 & 72.64 & 84.20 \\
& Ensemble & \textbf{49.51} & \textbf{66.02} & \textbf{76.36} & \textbf{91.10} \\
\bottomrule
\end{tabular}
\caption{Retrieval performance (Recall@1) on MSCOCO5k and Flickr30k at $256{\times}256$, comparing OpenLVD200M against Random Uniform Sampling.}
\label{tab:ablation_openlvd_ret}
\end{table}

\begin{table}[t]
\centering
\small
\resizebox{\columnwidth}{!}{
\begin{tabular}{lcccc}
\toprule
Benchmark & Metric & Random & OpenLVD200M & $\Delta$ \\
\midrule
FGVC-Aircraft & IT & 53.18 & \textbf{71.82} & \textbf{+18.64} \\
CUB-200                & IT & 70.40 & \textbf{73.95} & \textbf{+3.55} \\
ImageNet & (I-T)      & 70.51 & \textbf{73.74} & \textbf{+3.23} \\
ImageNet & (kNN)           & 76.18 & \textbf{78.07} & \textbf{+1.89} \\
\bottomrule
\vspace{-0.4cm}
\end{tabular}
}
\caption{OpenLVD200M: benchmark-specific improvements.\vspace{-0.4cm}}
\label{tab:openlvd_top3}
\end{table}

\section{Supplementary Results: Intermediate Model Sizes}
\label{sec:intermediate_models}

To explore the trade-off between model size and performance, we evaluate two intermediate-scale models distilled using the same recipe as the main SigLino-0.6B: SigLino-30m (30M parameters) and SigLino-70m (70M parameters). These models validate the scalability of our distillation approach across different parameter budgets and provide useful baselines for compute-constrained settings. Results show that even at significantly reduced model sizes, the multi-teacher distillation recipe maintains strong performance on classification, retrieval, and segmentation tasks.

\section{Details on OpenLVD200M Curation}
\label{sec:openlvd_details}

\begin{figure*}[t]
    \centering
    \includegraphics[width=0.9\textwidth]{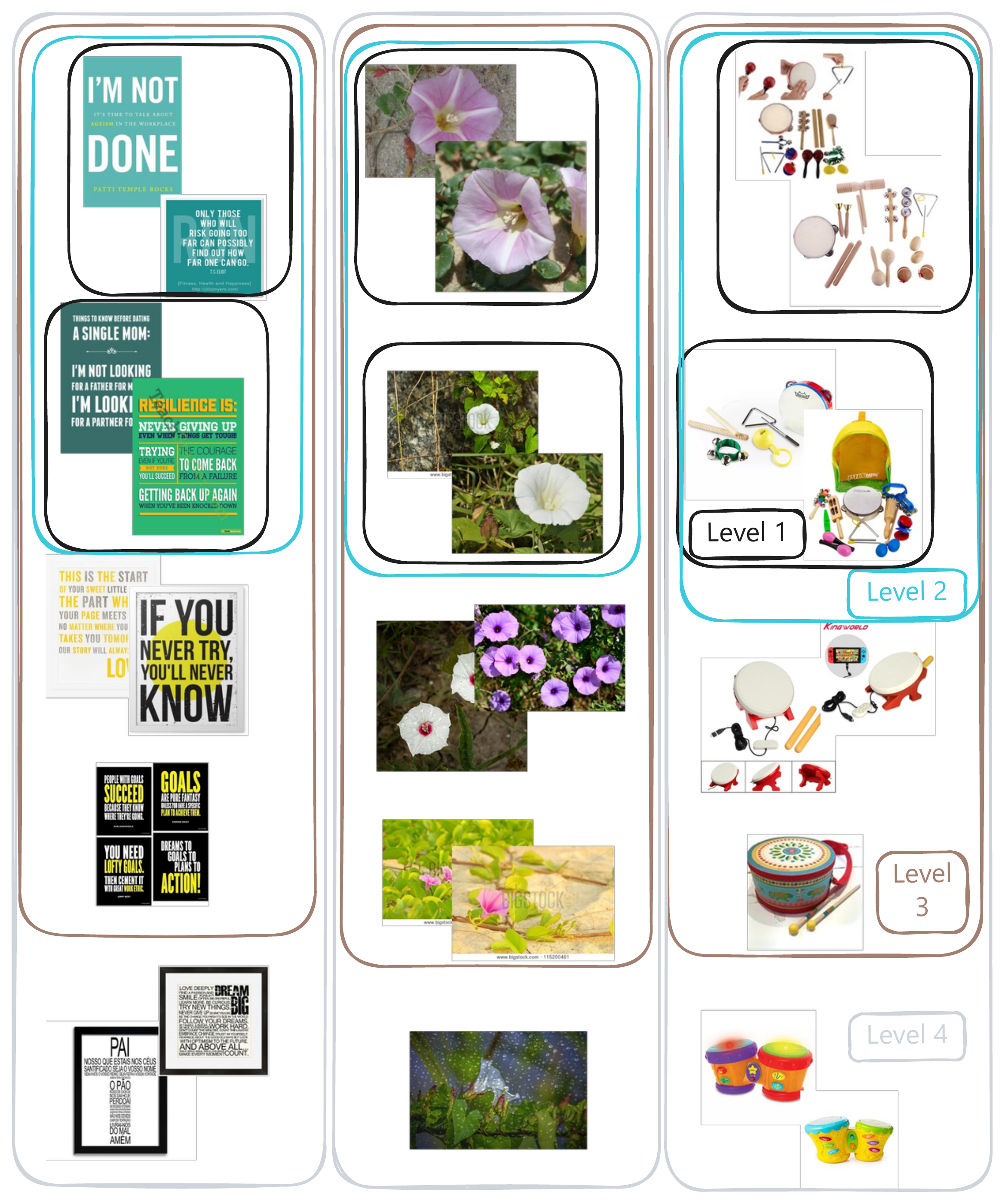}
    \caption{Concept hierarchy captured by the 4-level clustering. Each column represents a high-level semantic cluster (Level 4, grey borders), containing progressively finer granularities: Level 3 (brown borders), Level 2 (cyan borders), and Level 1 (black borders). From left to right, we show clusters for text-heavy images, flowers, and toys. The hierarchy naturally organizes concepts from broad categories to specific sub-types and fine-grained instances.}
    \label{fig:openlvd}
\end{figure*}

As outlined in \S\ref{sec:method}, we construct OpenLVD200M using the hierarchical clustering and sampling pipeline proposed by~\cite{vo2024automatic} to mitigate the long-tail biases inherent in web-scraped data. Figure~\ref{fig:openlvd} visually demonstrates the semantic structure captured by this process. The hierarchy organizes concepts from broad, high-level categories (Level 4, grey borders)—such as "text-heavy images", "flowers", or "musical instruments"—down to increasingly specific sub-types. By sampling uniformly across these nodes rather than the raw data distribution, we ensure that rare, fine-grained concepts (the leaves of the tree) are selected with the same probability as common head concepts. 

\paragraph{Implementation and Efficiency.} To scale this approach to our 2.3B image pool (DFN + LAION) using limited compute (12 nodes of 8$\times$A100), we introduce specific efficiency modifications to the original algorithm~\cite{vo2024automatic}. Instead of clustering the full dataset globally, we adopt a two-step assignment strategy:
(i) We embed all images using the DINOv3 ViT-B encoder.
(ii) We uniformly subsample a representative set of 1B images to learn the hierarchy via 4-level $k$-means, resulting in a tree structure with 20k (Level 4), 50k (Level 3), 500k (Level 2), and 20M (Level 1) centroids.
(iii) We assign the remaining 1.3B images to these pre-computed Level-1 centroids. 
(iv) We perform hierarchical sampling on the fully assigned population to produce the balanced 200M subset.

\begin{lstlisting}[language=Python, float=tp, caption={SigLino loss pseudo-code}]
def ComputeLoss(student, teachers, global_batch):
    L_total = 0
    # Gather global batch stats for stable normalization
    N_global = Sum(global_batch.num_images)
    For T in ["dino", "siglip"]:
        # Unpack per-image student (s) and teacher (t) features
        # s_sum/t_sum: Global Summary Token (CLS or Pooler)
        # s_pat/t_pat: Dense Patch Tokens
        s_sum, s_pat, s_reg = student[T]
        t_sum, t_pat, t_reg = teachers[T]

        # --- A. Local & Representation Alignment ---
        # Note: Patch loss normalized by token count per image (N_q)
        L_patch = Sum([MSE(s_pat[q], t_pat[q]) / N_q for q in batch])
        L_sum   = Sum([1 - CosineSim(s_sum[q], t_sum[q]) for q in batch])
        
        # DINOv3 specific: Align Registers
        if T == "dino":
             L_total += MSE(s_reg, t_reg)

        # --- B. ARKD ---        
        # 1. Compute Global Distance Matrices
        t_all = AllGather(t_sum)  # Gather from all ranks
        s_all = AllGather(s_sum)
        D_t = PairwiseDist(t_sum, t_all) # Teacher geometry
        D_s = PairwiseDist(s_sum, s_all) # Student geometry

        # 2. Normalize by Teacher Scale (Scale Invariance)
        scale = Mean(D_t)
        D_t, D_s = D_t / scale, D_s / scale

        # 3. Asymmetric Weighting (Intra-batch Median Split)
        median_dist = Median(D_t)
        # Penalize expansion only if samples are close (Intra-cluster)
        # Penalize shrinkage only if samples are far (Inter-cluster)
        W_expand = (D_t < median_dist)
        W_shrink = 1 - W_expand
        
        L_arkd = Mean(W_expand * SmoothL1(Max(D_s - D_t, 0)) + 
                      W_shrink * SmoothL1(Max(D_t - D_s, 0)))

        # Accumulate (Normalized by Global Batch Size)
        L_total += (L_patch + L_sum + L_arkd) / N_global

    return L_total
\end{lstlisting}
\label{lst:loss}

\end{document}